\documentclass[a4paper,twoside]{article}

\usepackage[dvipsnames,table]{xcolor}

\usepackage{epsfig}
\usepackage{subcaption}
\usepackage{calc}
\usepackage{amstext}
\usepackage{amsmath}
\usepackage{amsthm}
\usepackage{multicol}
\usepackage{pslatex}
\usepackage[bottom]{footmisc}

\usepackage{hyperref}
\usepackage{url}

\usepackage{natbib}
\usepackage{graphicx}
\usepackage[capitalize,nameinlink]{cleveref}
\usepackage{algorithm}
\usepackage{algorithmicx}
\usepackage[noend]{algpseudocode}
\usepackage{amssymb}
\usepackage{multirow}
\usepackage{tabularx}
\usepackage{tikz}
\usepackage{tikzfill}
\usepackage{stackengine}
\usepackage[nice]{nicefrac}
\usepackage{wrapfig}
\usepackage{booktabs}

\usepackage{mathtools}

\usepackage{amsmath,amsfonts,bm}

\def\eqref#1{equation~\ref{#1}}

\def\1{\bm{1}}

\DeclareMathAlphabet{\mathsfit}{\encodingdefault}{\sfdefault}{m}{sl}
\SetMathAlphabet{\mathsfit}{bold}{\encodingdefault}{\sfdefault}{bx}{n}

\newcommand{\Var}{\mathrm{Var}}

\newcommand{\newtext}[1]{#1}

\newcommand{\ie}{i.e.,\ }

\renewcommand{\eqref}[1]{(\ref{#1})}

\algblockdefx{Until}{EndUntil}[1]{\textbf{until} #1 \textbf{do}}{}
\algrenewcommand\algorithmicindent{4mm}
\algrenewcommand{\alglinenumber}[1]{\color{black!40}\fontsize{8.5}{8}\selectfont#1\color{black!30}:}
\definecolor{DarkGreen}{rgb}{0.0,0.6,0.0}
\newcommand{\AlgCommentTemplate}[2]{\hfill{\fontsize{8.5}{8}\selectfont\textcolor{DarkGreen}{\text{#1\;#2}}}}
\newcommand{\algCommentNoArrow}[1]{\AlgCommentTemplate{}{#1}}
\newcommand{\algComment}[1]{\AlgCommentTemplate{$\leftarrow$}{#1}}

\newcommand{\algCommentUp}[1]{
\AlgCommentTemplate{\rotatebox[origin=c]{90}{$\Rsh$}}{#1}}

\newcommand{\algCommentDown}[1]{\AlgCommentTemplate{\rotatebox[origin=c]{90}{$\Lsh$}}{#1}}

\algblockdefx{Forloop}{EndForloop}[1]{\textbf{for} #1 }{}
\algrenewcommand\algorithmicindent{4mm}

\newcommand{\dif}{\mathrm{d}}

\newcommand{\mymath}[2]{\newcommand{#1}{\TextOrMath{$#2$\xspace}{#2}}}
\newcommand{\totalloss}[1]{L_{#1}}
\newcommand{\totallossEst}[1]{\langle \nabla \totalloss{#1} \rangle}
\mymath{\dataset}{{\Omega}}
\mymath{\datasetSize}{{|\Omega|}}
\mymath{\batchSize}{{B}}
\mymath{\loss}{{\mathcal{L}}}
\mymath{\lossce}{\loss_\text{cross-ent}}
\mymath{\score}{{s}}
\mymath{\target}{y}
\mymath{\outputlayer}{m(x_i,\theta)}
\mymath{\memory}{q}

\mymath{\model}{m}
\mymath{\momentum}{\alpha}

\mymath{\unitfunc}{\mathbf{1}}
\mymath{\classcount}{J}
\mymath{\weight}{w}
\mymath{\pdf}{p}

\DeclareGraphicsExtensions{.png,.jpg,.pdf,.ai,.psd}
\usepackage{SCITEPRESS}     %

\begin{document}

\title{Multiple Importance Sampling for Stochastic Gradient Estimation}

\author{
    \authorname{
        Corentin Salaün\sup{1}\orcidAuthor{0000-0002-5112-7488}
        ~Xingchang Huang\sup{1}\orcidAuthor{0000-0002-2769-8408}
        ~Iliyan Georgiev\sup{2}\orcidAuthor{0000-0002-9655-2138}
        ~Niloy Mitra\sup{2,3}\orcidAuthor{0000-0002-2597-0914}
        ~Gurprit Singh\sup{1}\orcidAuthor{0000-0003-0970-5835}
    }
    \affiliation{\sup{1}Max Planck Institute for Informatics, Saarland University, Saarbrücken, Germany}
    \affiliation{\sup{2}Adobe Research, London, United Kingdom}
    \affiliation{\sup{3}Department of Computer Science, University College London, London, United Kingdom}
    \email{\{csalaun,xhuang,gsingh\}@mpi-inf.mpg.de, igeorgiev@adobe.com, n.mitra@cs.ucl.ac.uk}
}


\abstract{We introduce a theoretical and practical framework for efficient importance sampling of mini-batch samples for gradient estimation from single and multiple probability distributions. 
To handle noisy gradients, our framework dynamically evolves the importance distribution during training by utilizing a self-adaptive metric.  
Our framework combines multiple, diverse sampling distributions, each tailored to specific parameter gradients.
This approach facilitates the importance sampling of \emph{vector-valued} gradient estimation. 
Rather than naively combining multiple distributions, our framework involves optimally weighting data contribution across multiple distributions. This adapted combination of multiple importance yields superior gradient estimates, leading to faster training convergence.
We demonstrate the effectiveness of our approach through empirical evaluations across a range of optimization tasks like classification and regression on both image and point cloud datasets.
}

\onecolumn \maketitle \normalsize \setcounter{footnote}{0} \vfill

\section{Introduction}

Stochastic gradient descent (SGD) is fundamental in optimizing complex neural networks. This iterative optimization process relies on the efficient estimation of gradients to update model parameters and minimize the optimization objective. A significant challenge in methods based on SGD lies in the influence of stochasticity on gradient estimation, impacting both the quality of the estimates and convergence speed. This stochasticity introduces errors in the form of noise, and addressing and minimizing such noise in gradient estimation continues to be an active area of research.

Various approaches have been introduced to reduce gradient estimation noise, including data diversification \cite{zhang2019active,faghri2020study,ren2019adaptive}, adaptive mini-batch sizes~\cite{balles2017coupling,alfarra2021adaptive}, momentum-based estimation \cite{rumelhart1986learning,kingma2014adam}, and adaptive sampling strategies \cite{santiago2021low}. These methods collectively expedite the optimization by improving the gradient-estimation accuracy.

Another well-established technique for noise reduction in estimation is importance sampling (IS)~\cite{loshchilov2015online,katharopoulos2017biased,katharopoulos2018dlis}, which involves the non-uniform selection of data samples for mini-batch construction. Data samples that contribute more significantly to gradient estimation are selected more often. This allows computational resources to focus on the most critical data for the optimization task.
However, these algorithms are quite inefficient and add significant overhead to the training process.
Another limitation of importance sampling, in general, lies in determining the best sampling distribution to achieve maximal improvement, often necessitating a quality trade-off due to the simultaneous estimation of numerous parameters. 

We propose an efficient importance sampling algorithm that does \textit{not} require resampling, in contrast to \cite{katharopoulos2018dlis}. 
Our importance function dynamically evolves during training, utilizing a self-adaptive metric to
effectively manage initial noisy gradients. 
Further, unlike existing IS methods in machine learning where importance distributions assume scalar-valued gradients, we propose a multiple importance sampling (MIS) strategy to manage \emph{vector-valued} gradient estimation \newtext{(\ie multiple parameters)}.
We propose the simultaneous use of multiple sampling strategies combined with a weighting approach following the principles of MIS theory, well studied in the rendering literature in computer graphics~\cite{veach1997robust}.
Rather than naively combining multiple distributions, our proposal involves estimating importance weights w.r.t.\ data samples
across multiple distributions by leveraging the theory of optimal MIS (OMIS)~\cite{OptiMIS}. 
This optimization process yields superior gradient estimates, leading to faster training convergence. 
In summary, we make the following contributions:
\begin{itemize}%
\item An efficient IS algorithm with a self-adaptive metric for importance sampling is developed.
\item An MIS estimator for gradient estimation is introduced to improve gradients estimation.
\item A practical approach to computing the OMIS weights is presented to maximize the quality of vector-valued gradient estimation.
\item The effectiveness of the approach is demonstrated on various machine learning tasks.
\end{itemize}

\section{Related work}

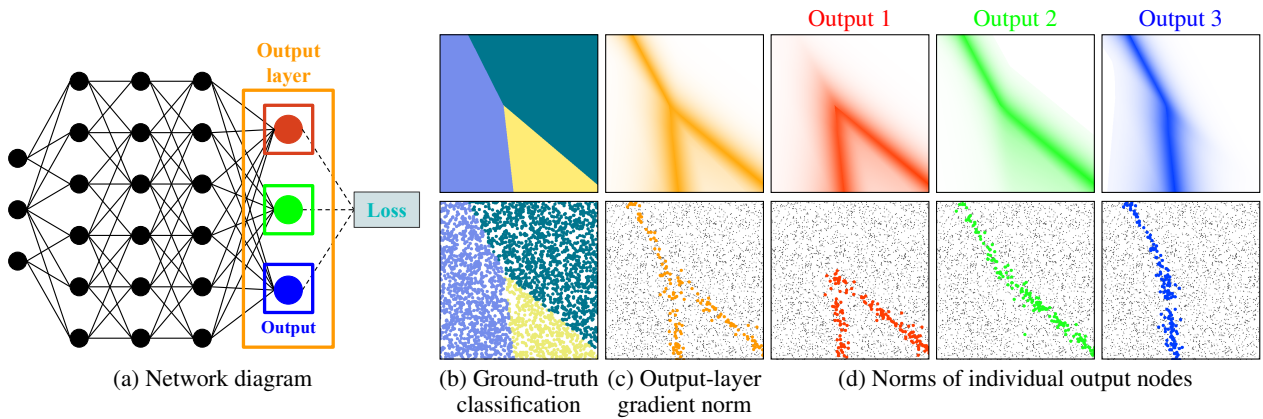
\begin{figure*}[t]
    \centering
    \resizebox{0.9\textwidth}{!}{
    \newcommand{\PlotSingleImage}[1]{%
        \begin{scope}
            \clip (0,0) -- (2.5,0) -- (2.5,2.5) -- (0,2.5) -- cycle;
            \path[fill overzoom image=#1] (0,0) rectangle (2.5cm,2.5cm);
        \end{scope}
        \draw (0,0) -- (2.5,0) -- (2.5,2.5) -- (0,2.5) -- cycle;
}
\newcommand{\PlotSchema}[1]{%
        \begin{scope}
            \path[fill overzoom image=#1] (0,0) rectangle (3.4cm,2.5cm);
        \end{scope}

}

\newcommand{\scaleval}{1.72}
\small
\hspace*{-3mm}
\begin{tabular}{c@{\;}c@{\;}c@{\;}c@{\;}c@{\;}c@{}}
~ & ~ & & \footnotesize \textcolor[rgb]{1,0,0}{Output 1} & \footnotesize \textcolor[rgb]{0,1,0}{Output 2} & \footnotesize \textcolor[rgb]{0,0,1}{Output 3} \\
\multirow{2}{*}[0.72in]{\begin{tikzpicture}[scale=\scaleval]    
     \PlotSchema{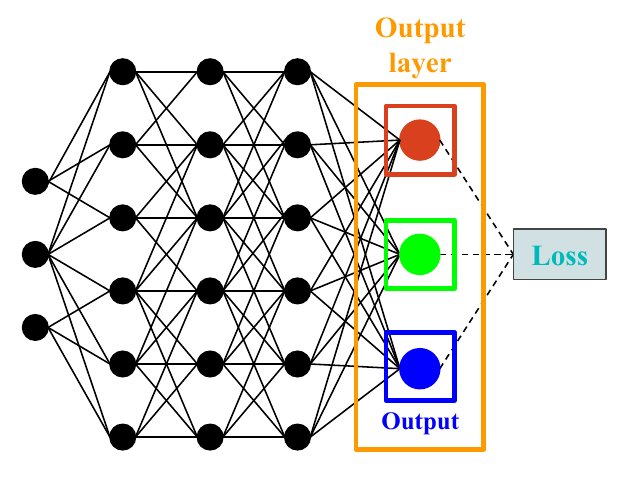}
\end{tikzpicture}}
&
\begin{tikzpicture}[scale=\scaleval/2.05]    
     \PlotSingleImage{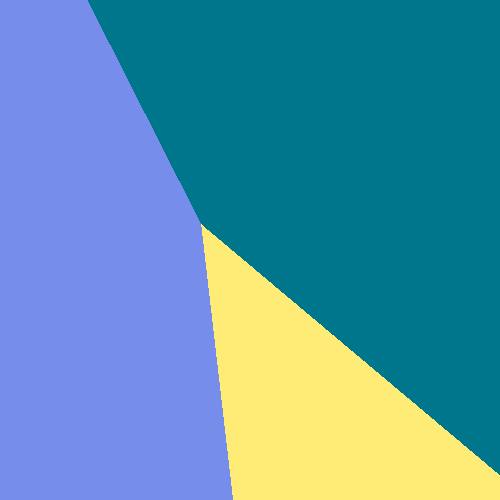}
\end{tikzpicture}
&
\begin{tikzpicture}[scale=\scaleval/2.05]    
     \PlotSingleImage{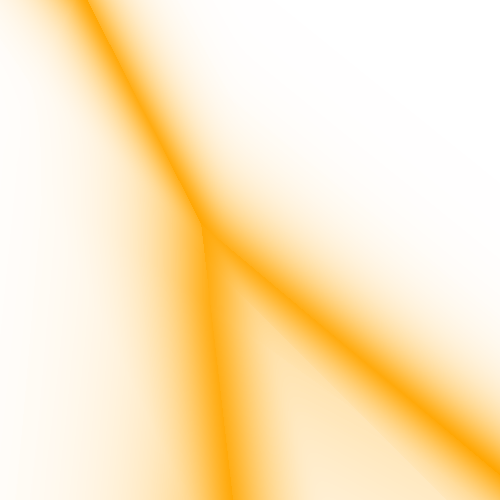}
\end{tikzpicture}
&
\begin{tikzpicture}[scale=\scaleval/2.05]    
    \PlotSingleImage{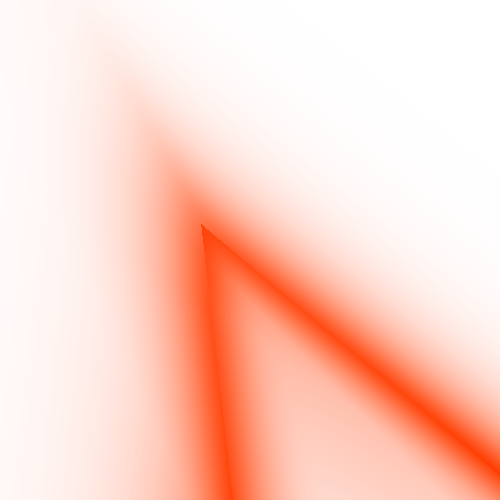}
\end{tikzpicture}
&
\begin{tikzpicture}[scale=\scaleval/2.05]    
    \PlotSingleImage{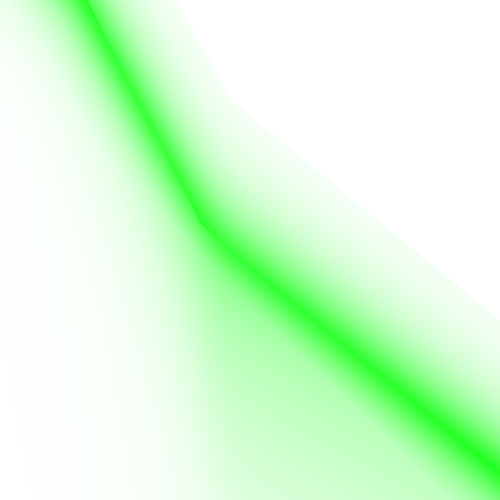}
\end{tikzpicture}
&
\begin{tikzpicture}[scale=\scaleval/2.05]    
    \PlotSingleImage{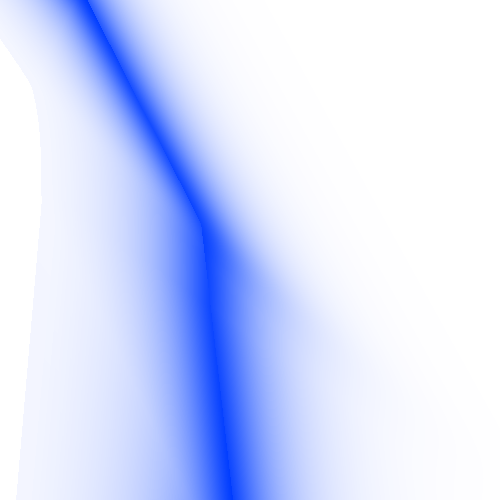}
\end{tikzpicture}
\\
~
&
\begin{tikzpicture}[scale=\scaleval/2.05]    
     \PlotSingleImage{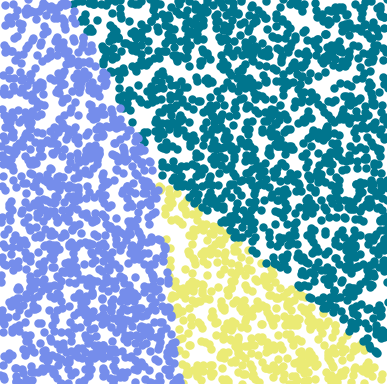}
\end{tikzpicture}
&
\begin{tikzpicture}[scale=\scaleval/2.05]    
     \PlotSingleImage{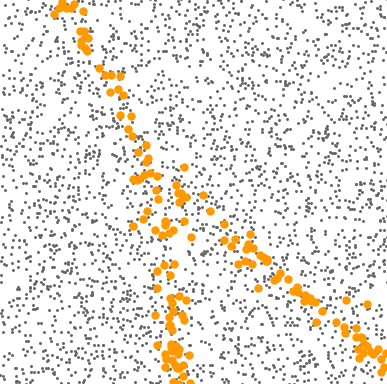}
\end{tikzpicture}
&
\begin{tikzpicture}[scale=\scaleval/2.05]    
     \PlotSingleImage{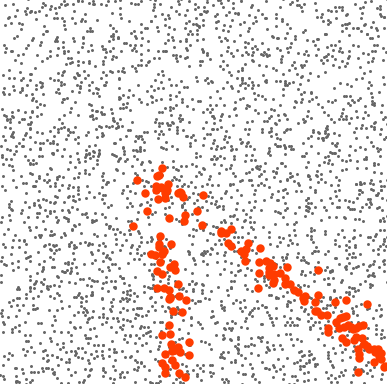}
\end{tikzpicture}
&
\begin{tikzpicture}[scale=\scaleval/2.05]    
    \PlotSingleImage{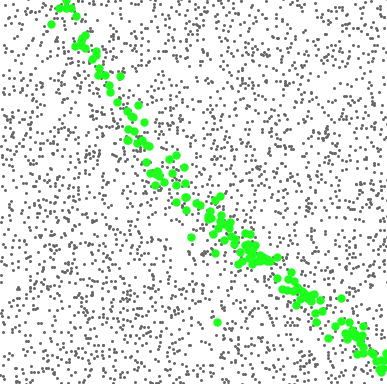}
\end{tikzpicture}
&
\begin{tikzpicture}[scale=\scaleval/2.05]    
    \PlotSingleImage{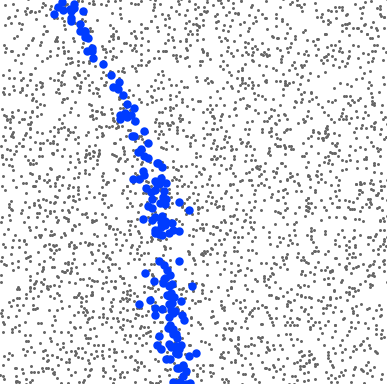}
\end{tikzpicture}
\\%[2mm]
\footnotesize (a) Network diagram & \footnotesize (b) Ground-truth & (c) Output-layer & \multicolumn{3}{c}{\footnotesize (d) Norms of individual output nodes} \\
& \footnotesize classification & gradient norm & 
\end{tabular}
    }
    \vspace{1mm}
    \caption{
        We visualize different importance sampling distributions for a simple classification task.
        We propose to use the output layer gradients for importance sampling, as shown in the network diagram (a).
        For a given ground-truth classification (top) and training dataset (bottom) shown in (b), it is possible to importance sample from the $L_2$ norm of the output-layer gradients (c) or from 
        three different sampling distributions derived from the gradient norms of individual output nodes (d). 
        The bottom row shows sample weights from each distribution.
    }
    \label{fig:2D_Sampling_visualization}
\end{figure*}

\paragraph{Importance sampling for gradient estimation.}

Importance sampling (IS)~\cite{kahn1950random,kahn1953methods,owen2000safe} has emerged as a powerful technique in high energy physics, Bayesian
inference, rare event simulation for finance and insurance, and rendering in
computer graphics. 
In the past few years, IS has also been applied in machine learning to improve the accuracy of gradient estimation and enhance the overall performance of learning algorithms~\cite{zhaoa2015stochastic}. 

By strategically sampling data points from a non-uniform distribution, IS effectively focuses training resources on the most informative and impactful data, leading to more accurate gradient estimates.
\citet{bordes2005fast} developed an online algorithm (LASVM) that uses importance sampling to train kernelized support vector machines.
\citet{loshchilov2015online} suggested employing data rankings based on their respective loss values. This ranking is then employed to create an importance sampling strategy that assigns greater importance to data with higher loss values.
\citet{katharopoulos2017biased} proposed importance sampling the loss function. Subsequently, \citet{katharopoulos2018dlis} introduced an upper bound to the gradient norm that can be employed as an importance function. Their algorithm involves resampling and computing gradients with respect to the final layer. Despite the importance function demonstrating improvement over uniform sampling, their algorithm exhibits significant inefficiency.

\paragraph{Multiple importance sampling.}

The concept of Multiple Importance Sampling (MIS) emerged as a robust and efficient technique for integrating multiple sampling strategies~\cite{owen2000safe}. Its core principle lies in assigning weights to \newtext{multiple importance sampling estimator, each using a different sampling distribution}, allowing each data sample to utilize the most appropriate strategy. 
\Citet{veach1997robust} introduced this concept of MIS to rendering in computer graphics and proposed the widely adopted \emph{balance heuristic} for importance (weight) allocation. 
The balance heuristic determines weights based on a data sample's relative importance across all sampling approaches, effectively mitigating the influence of outliers with low probability densities. While MIS is straightforward to implement and independent of the specific function, Variance-Aware MIS \cite{grittmann2019Variance} advanced the concept by using variance estimates from each sampling technique for further error reduction. Moreover, Optimal MIS \cite{OptiMIS} derived optimal sampling weights that minimize MIS estimator variance. Notably, these weights depend not only on probability density but also on the function values of the samples. \Cref{app:MIS} summarizes the theory behind (multiple) importance sampling. It also states the optimal MIS estimator and how to compute it.

\section{Problem statement}

The primary goal of machine-learning optimization is to find the optimal parameters $\theta$ for a given model function $\model(x, \theta)$ by minimizing a loss function $\loss$ over a dataset $\dataset$:
\begin{align}
    \label{eq:ProblemStatement}
    \theta^{*} = \underset{\theta}{\mathrm{argmin}} \, \underbrace{\int_{\dataset} \loss(\outputlayer,y) \, \dif x.}_{\totalloss{\theta}}
\end{align}
The loss function $\loss$ quantifies the dissimilarity between the model predictions $\model(x,\theta)$ and observed data $y$. In the common case of a discrete dataset, the integral becomes a sum.

In practice, the total loss is minimized via iterative gradient descent. In each iteration $t$, the gradient $\nabla \totalloss{\theta_t}$ of the loss with respect to the current model parameters $\theta_t$ is computed, and the parameters are updated as
\begin{equation}
    \label{eq:gradientDescent}
    \theta_{t+1} = \theta_t - \lambda \underbrace{\int_{\dataset} \nabla \loss(\model(x,\theta_t),y) \, \dif x}_{\nabla \totalloss{\theta_t}},
\end{equation}
\newtext{where $\lambda > 0$ is the learning rate. It is also possible to use an adaptive learning rate instead of a constant.}

\paragraph{Monte Carlo gradient estimator.}

In practice, the parameter gradient is estimated from a small batch $\{ x_i \}_{i=1}^\batchSize$ of randomly selected data points:
\begin{multline}
    \totallossEst{\theta} = \sum_{i=1}^\batchSize \frac{\nabla \loss(\model(x_i,\theta),y_i)}{\batchSize \pdf(x_i)} \approx \nabla \totalloss{\theta},
    \quad
    x_i \sim \pdf.
    \label{eq:MC_Gradient}
\end{multline}
The data points are sampled from a probability density function (pdf) $\pdf$ or probability mass function in discrete cases. 
The mini-batch gradient descent substitutes the true gradient $\nabla \totalloss{\theta_t}$ with an estimate $\totallossEst{\theta_t}$ in \cref{eq:gradientDescent} to update the model parameters in each iteration.

We want to estimate $\nabla \totalloss{\theta_t}$ accurately and also efficiently, since the gradient-descent iteration~\eqref{eq:gradientDescent} may require many thousands of iterations until the parameters converge.
These goals can be achieved by performing the optimization in small batches whose samples are chosen according to a carefully designed distribution $p$.
For a simple classification problem, \cref{fig:2D_Sampling_visualization}c shows an example importance sampling distribution derived from the output layer of the model.
In \cref{fig:2D_Sampling_visualization}d we derive multiple distributions from the individual output nodes.
Below we develop theory and practical algorithms for importance sampling using a single distribution (\cref{sec:importance_sampling}) and for combining multiple distributions to further improve gradient estimation (\cref{sec:multiple_importance_sampling}).

\section{Mini-batch importance sampling}
\label{sec:importance_sampling}

Mini-batch gradient estimation~\eqref{eq:MC_Gradient} notoriously suffers from Monte Carlo noise, which can make the parameter-optimization trajectory erratic and convergence slow. That noise comes from the often vastly different contributions of different samples $x_i$ to that estimate.

Typically, the selection of \newtext{the multiple samples constructing a mini-batch is done with uniform probability $\pdf(x_i)=1/\datasetSize$. Each data of the mini-batch is sampled with replacement following this distribution}. Importance sampling is a technique for using a non-uniform pdf to strategically pick samples proportionally on their contribution to the gradient, to reduce estimation variance.

\paragraph{Practical algorithm.}

We propose an importance sampling algorithm for mini-batch gradient descent, outlined in~\cref{alg:IS}. Similarly to~\citet{schaul2015prioritized}, we use an importance function that relies on readily available quantities for each data point, introducing only negligible memory and computational overhead over classical uniform mini-batching. We store a set of persistent \emph{un-normalized importance} scalars $\memory = \{ \memory_i \}_{i=1}^\datasetSize$ that are updated continuously during the optimization. 

The first epoch is a standard SGD one, during which we additionally compute the initial importance of each data point (line 3).
In each subsequent epoch, at each mini-batch optimization step $t$ we normalize the importance values to a valid distribution $p$ (line 6). We then choose $\batchSize$ data samples (with replacement) according to $p$ (line 7). The loss $\loss$ is evaluated for each selected data sample (line 8), and backpropagated to compute the loss gradient (line 9). The per-sample importance is used in the gradient estimation (line 10) to normalize the contribution. In practice lines 9-10 can be done simultaneously by backpropagating a weighted loss $\loss(x) \cdot (\nicefrac{1}{(p(x)\cdot B)})^{T}$. Finally, the network parameters are updated using the estimated gradient (line 11). On line 12, we update the importance of the samples in the mini-batch; we describe our choice of importance function below. The blending parameter $\momentum$ ensures stability of the persistent importance as discussed in~\cref{sec:importance_momentum}. At the end of each epoch (line 14), we add a small value to the un-normalized weights of all data to ensure that every data point will be eventually evaluated, even if its importance is deemed low by the importance metric.

\begin{algorithm}[t]
    \caption{Mini-batch importance sampling for SGD.}
    \small
    \begin{algorithmic}[1]
        \State $\theta \leftarrow $ random parameter initialization
        \State $B \leftarrow$ mini-batch size, $N=|\Omega|$ \algComment{Dataset size}
        \State $q,\theta \leftarrow \text{Initialize}(x,y,\Omega,\theta,B)$ 
        \algComment{\cref{alg:Initialization}}
        \Until{convergence} \algComment{Loop over epochs}
            \Forloop{$t \leftarrow 1$ \textbf{to} $N/B$} \algComment{Loop over mini-batches}
                \State $p \leftarrow \memory/$sum$(\memory)$ \algComment{Normalize importance to pdf}
                \State $x,y \leftarrow \batchSize$ data samples $\{x_i, y_i \}_{i=1}^{B} \propto p$
                \State $\loss(x) \leftarrow \loss(\model(x,\theta),y)$
                \State $\nabla \loss(x) \leftarrow $ Backpropagate$(\loss(x)) $
                \State $\totallossEst{\theta} \leftarrow (\nabla \loss(x) \cdot (\nicefrac{1}{p(x)})^{T})/B$ \algComment{\cref{eq:MC_Gradient}}
                \State $\theta \leftarrow \theta - \lambda \, \totallossEst{\theta}$ \algComment{SGD step}
                \State $\memory(x) \leftarrow  \momentum\cdot\memory(x) + (1-\momentum)\cdot\left\Vert \frac{\partial \loss(x)}{\partial \model(x,\theta)} \right\Vert$
            \EndForloop
            \State $\memory \leftarrow \memory + \epsilon$ \algCommentUp{Accumulate importance}
        \EndUntil
        \Return $\theta$ 
    \end{algorithmic}
    \label{alg:IS}
\end{algorithm}

It is important to note that the first epoch is done without importance sampling to initialize each sample importance. This does not add overhead as it is equivalent to a classical epoch running over all data samples. While similar schemes have been proposed in the past~\cite{loshchilov2015online}, they often rely on a multitude of hyperparameters, making their practical implementation challenging. This has led to the development of alternative methods like re-sampling \cite{katharopoulos2018dlis,dong2021one,zhang2023adaselection}. Tracking importance across batches and epochs minimizes the computational overhead, further enhancing the efficiency and practicality of the approach.

\paragraph{Importance function.}

In combination with the presented algorithm, we propose an importance function that is efficient to evaluate. While the gradient $L_2$ norm has been shown to be optimal \cite{zhaoa2015stochastic,needell2014stochastic,wang2017accelerating,alain2015variance}, calculating it can be computationally expensive as it requires full backpropagation for every data point. To this end, we compute the gradient norm only for a subset of the parameters, specifically the output nodes of the network: $\memory(x) = \left\Vert \frac{\partial \mathcal{L}(x)}{\partial \model(x,\theta)} \right\Vert$. This choice is based on an upper bound of the gradient norm, using the chain rule and the Cauchy–Schwarz inequality \cite{katharopoulos2018dlis}:
\begin{align}
    \label{eq:GradNormBound}
    \!\left\Vert \frac{\partial \mathcal{L}(x_i)}{\partial \theta} \right\Vert
        &= \left\Vert \frac{\partial \mathcal{L}(x)}{\partial \model(x,\theta)} \cdot \frac{\partial \model(x,\theta)}{\partial \theta} \right\Vert \leq \\
        &\left\Vert \frac{\partial \mathcal{L}(x)}{\partial \model(x,\theta)} \right\Vert \cdot \left\Vert \frac{\partial \model(x,\theta)}{\partial \theta} \right\Vert \leq \underbrace{\left\Vert \frac{\partial \mathcal{L}(x)}{\partial \model(x,\theta)} \right\Vert}_{\memory(x)} \cdot \, C \, , \nonumber
\end{align}
where $C$ is the Lipschitz constant of the parameters gradient. That is, our importance function is a bound of the gradient magnitude based on the output-layer gradient norm. 

We tested the relationship between four different importance distributions: uniform, our proposed importance function,
\begin{wrapfigure}{r}{0.25\textwidth}
    \vspace{-10pt}
    \includegraphics[width=0.25\textwidth]{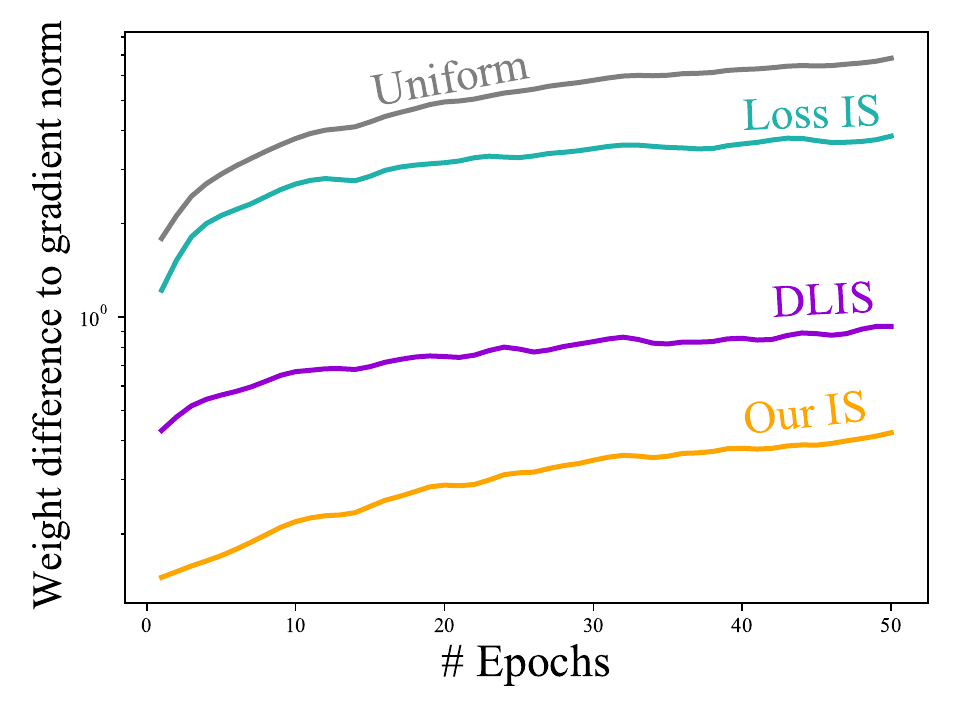}
    \vspace{-25pt}
\end{wrapfigure}
the loss function as importance~\cite{katharopoulos2017biased}, and the work by \citet{katharopoulos2018dlis} using an other gradient norm bound. The inline figure plots the $L_2$ difference between these importance distributions and the ground-truth gradient-norm distribution across epochs for an MNIST classification task. It shows that \textcolor[rgb]{1,0.7,0}{Our IS} distribution has the smallest difference, i.e., it achieves high accuracy while requiring only a small part of the gradient. 

For some specific task when the output layer has predictable shape, it is possible to derive a closed form definition of the proposed importance metric. \Cref{app:Cross_entropy_close_form} derives the close form importance for classification task using cross entropy loss.

Note that any importance heuristic can be used on line 12 of \cref{alg:IS}, such as the gradient norm~\cite{zhaoa2015stochastic,needell2014stochastic,wang2017accelerating,alain2015variance}, the loss~\cite{loshchilov2015online,katharopoulos2017biased,dong2021one}, or more advanced importance~\cite{katharopoulos2018dlis}. For efficiency, our importance function reuses the forward-pass computations from line 8, updating $q$ only for the current mini-batch samples. 

\section{Multiple importance sampling}
\label{sec:multiple_importance_sampling}

The parameter gradient $\nabla\totalloss{\theta}$ is vector with dimension equal to the number of model parameters. The individual parameter derivatives vary uniquely across the data points, and estimation using a single distribution (\cref{sec:importance_sampling}) inevitably requires making a trade-off, e.g., only importance sampling the overall gradient magnitude. Truly minimizing the estimation error requires estimating each derivative using a separate importance sampling distribution tailored to its variation. However, there are two practical issues with this approach: First, it would necessitate sampling from all of these distributions, requiring ``mini-batches'' of size equal at least to the number of parameters. Second, it would lead to significant computation waste, since backpropagation computes all parameter derivatives but only one of them would be used per data sample. To address this issue, we propose using a small number of distributions, each tailored to the variation of a parameter subset, and combining \emph{all} computed derivatives into a low-variance estimator, using multiple importance sampling theory. As an example, \cref{fig:2D_Sampling_visualization}d shows three sampling distributions for a simple classification task, based on the derivatives of the network's output nodes, following the boundary of each class. 

\paragraph{MIS gradient estimator.}

Combining multiple sampling distributions into a single robust estimator has been well studied in the Monte Carlo rendering literature. The best known method is \emph{multiple importance sampling} (MIS)~\cite{veach1997robust}. In our case of gradient estimation, the MIS estimator takes for form
\begin{equation}
    \label{eq:MIS_Monte_Carlo_estimator}
    \totallossEst{\theta}_\mathrm{MIS} = \sum_{j=1}^{J}\sum_{i=1}^{n_j}w_{j}(x_{ij})\frac{\nabla \loss(\model(x_{ij},\theta),y_{ij})}{n_jp_j(x_{ij})},
\end{equation} 
where $J$ is the number of sampling distributions, $n_j$ the number of samples from distribution $j$, and $x_{ij}$ the $i$\textsuperscript{th} sample from the $j$\textsuperscript{th} distribution. Each sample is modulated by a weight $w_{j}(x_{ij})$; the estimator is unbiased as long as $\sum_{j=1}^J w_j(x) = 1$ for every data point $x$ in the dataset.

\paragraph{Optimal weighting.}
\label{sec:optimal_mis}

Various MIS weighting functions $w_j$ have been proposed in literature, the most universally used one being the balance heuristic~\cite{veach1997robust}. In this work we use the recently derived optimal weighting scheme~\cite{OptiMIS} which minimizes the estimation variance for a given set of sampling \mbox{distributions $p_j$:}
\begin{multline}
    \label{eq:OptiMISweigth}
    w_{j}(x) = \alpha_j \frac{p_j(x)}{\nabla \loss(\model(x,\theta),y)} \; + \\
    \frac{n_jp_j(x)}{\sum_{k=1}^J n_kp_k(x)}\Bigg(1 - \frac{\sum_{k=1}^J \alpha_kp_k(x)}{\nabla \loss(\model(x,\theta),y)}\Bigg).
\end{multline}
Here, $\boldsymbol{\alpha} = [ \alpha_1, \ldots, \alpha_J ]$ is the solution to the linear system
\begin{equation}
    \label{eq:AlphaDefinition}
    \small
    \boldsymbol{A} \boldsymbol{\alpha} = \boldsymbol{b} \text{, with }
    \begin{dcases}
        a_{j,k} = \int_{\dataset} \frac{\pdf_j(x) \pdf_k(x)}{\sum_{i}^{J} n_i p_i(x)}d(x,y),\\
        b_j = \int_{\dataset} \frac{\pdf_j(x) \nabla \loss(\model(x,\theta),y)}{\sum_{i}^{J} n_i p_i(x)}d(x,y),
    \end{dcases}
\end{equation}
where $a_{j,k}$ and $b_j$ are the elements of the matrix $\boldsymbol{A} \in \mathbb{R}^{J \times J}$ and vector $\boldsymbol{b} \in \mathbb{R}^J$ respectively.

Instead of explicitly computing the optimal weights in \cref{eq:OptiMISweigth} using \cref{eq:AlphaDefinition} and plugging them into the MIS estimator \eqref{eq:MIS_Monte_Carlo_estimator}, we can use a shortcut evaluation that yields the same result~\cite{OptiMIS}:
\begin{equation}
    \label{eq:OMIS}
    \totallossEst{\theta}_\mathrm{OMIS} = \sum_{j = 1}^J \alpha_j.
\end{equation}

In \cref{app:MIS} we provide an overview of MIS and the aforementioned weighting schemes. Importantly for our case, the widely adopted balance heuristic does not bring practical advantage over single-distribution importance sampling  (\cref{sec:importance_sampling}) as it is equivalent to sampling from a mixture of the given distributions; we can easily sample from this mixture by explicitly averaging the distributions into a single one. In contrast, the optimal weights are different for each gradient dimension as they depend on the gradient value.

\begin{algorithm}[t]
    \caption{
        Optimal multiple importance sampling SGD.
    }
    \label{alg:OMIS}
    \small
    \begin{algorithmic}[1]
        \State $\theta \leftarrow $ random parameter initialization
        \State $B \leftarrow$ mini-batch size, $J \leftarrow$ number of pdf
        \State $N=|\Omega| \leftarrow $ dataset size
        \State $n_j \leftarrow $ sample count per technique, for $j \in \{1,..J\}$
        \State $\boldsymbol{q},\theta \leftarrow \text{InitializeMIS}(x,y,\Omega,\theta,B)$ \algComment{\cref{alg:Initialization_MIS}}
        \State $\langle\boldsymbol{A} \rangle \leftarrow 0^{J \times J} , \langle\boldsymbol{b} \rangle \leftarrow 0^{J}$ \algComment{OMIS linear system}
        \Until{convergence} \algComment{Loop over epochs}
            \Forloop{$t \leftarrow 1$ \textbf{to} $N/B$} \algComment{Loop over mini-batches}
                \State $\langle\boldsymbol{A} \rangle \leftarrow \beta\langle\boldsymbol{A} \rangle , \langle\boldsymbol{b} \rangle \leftarrow \beta\langle\boldsymbol{b} \rangle$
                \Forloop{$j \leftarrow 1$ \textbf{to} $J$} \algComment{Loop over distributions}
                    \State $p_j \leftarrow \memory_j/\text{sum}(\memory_j)$
                    \State $x,y \leftarrow \batchSize$ data samples $\{x_i, y_i \}_{i=1}^{n_j} \propto p_j$
                    \State $\loss(x) \leftarrow \loss(\model(x,\theta),y)$
                    \State $\nabla \loss(x) \leftarrow $ Backpropagate$(\loss(x))$
                    \State $S(x) \leftarrow \sum_{k=1}^J n_k \pdf_k(x)$
                    \State $\boldsymbol{W} \leftarrow \nicefrac{n_i\pdf_i(x)}{\sum_{k=1}^J n_k \pdf_k(x)}$ \algCommentDown{Momentum estim.}
                    \State $\langle\boldsymbol{A}\rangle \leftarrow \langle\boldsymbol{A}\rangle  + (1-\beta)\sum_{i = 1}^{n_j} \boldsymbol{W}_{i}\boldsymbol{W}_{i}^T$
                    \State $\langle \boldsymbol{b} \rangle \leftarrow \langle\boldsymbol{b} \rangle  + (1-\beta)\sum_{i = 1}^{n_j} \nabla \loss(x_i)\nicefrac{\boldsymbol{W}_{i}}{S(x_i)}$
                    \State $ \boldsymbol{\memory}(x) \leftarrow \momentum \boldsymbol{\memory}(x) + (1-\momentum) \frac{\partial \mathcal{L}(x)}{\partial \model(x,\theta)}$ %
                \EndForloop
                \State $ \langle \boldsymbol{\alpha} \rangle \leftarrow \langle \boldsymbol{A} \rangle^{-1}\langle \boldsymbol{b} \rangle$
                \State $ \totallossEst{\theta}_\mathrm{OMIS} \leftarrow \sum_{j = 1}^J \langle \alpha_j \rangle$
                \State $\theta \leftarrow \theta - \eta \, \totallossEst{\theta}_\mathrm{OMIS}$ \algComment{SGD step}
            \EndForloop
        \EndUntil
        \Return $\theta$
    \end{algorithmic}
\end{algorithm}

\paragraph{Practical algorithm.}

Implementing the optimal-MIS estimator~\eqref{eq:OMIS} amounts to drawing $n_j$ samples from each distribution, computing $\boldsymbol{\alpha}$ for each dimension of the gradient and summing its elements. The integrals in $\boldsymbol{A}$ and $\boldsymbol{b}$ (sums in the discrete-dataset case) can be estimated as $\langle\boldsymbol{A}\rangle$ and $\langle\boldsymbol{b}\rangle$ from the drawn samples, yielding the estimate $\langle\boldsymbol{\alpha}\rangle = \langle\boldsymbol{A}\rangle^{-1} \langle\boldsymbol{b}\rangle$.

\Cref{alg:OMIS} shows a complete gradient-descent algorithm. The main differences with \cref{alg:IS} are the use of multiple importance distributions $\boldsymbol{\memory} = \{q_j\}_{j=1}^J$ (line 5) and the linear system used to compute the OMIS estimator (line 6). This linear system is updated (lines 15-18) using the mini-batch samples and solved to obtain the gradient estimation (line 22). Since the matrix $\langle\boldsymbol{A}\rangle$ is independent of the gradient estimation (see \cref{eq:AlphaDefinition}), its inversion can be shared across all parameter estimates.

\begin{figure}[t]
    \centering
    \resizebox{0.9\columnwidth}{!}{
    \includegraphics[width=0.45\textwidth]{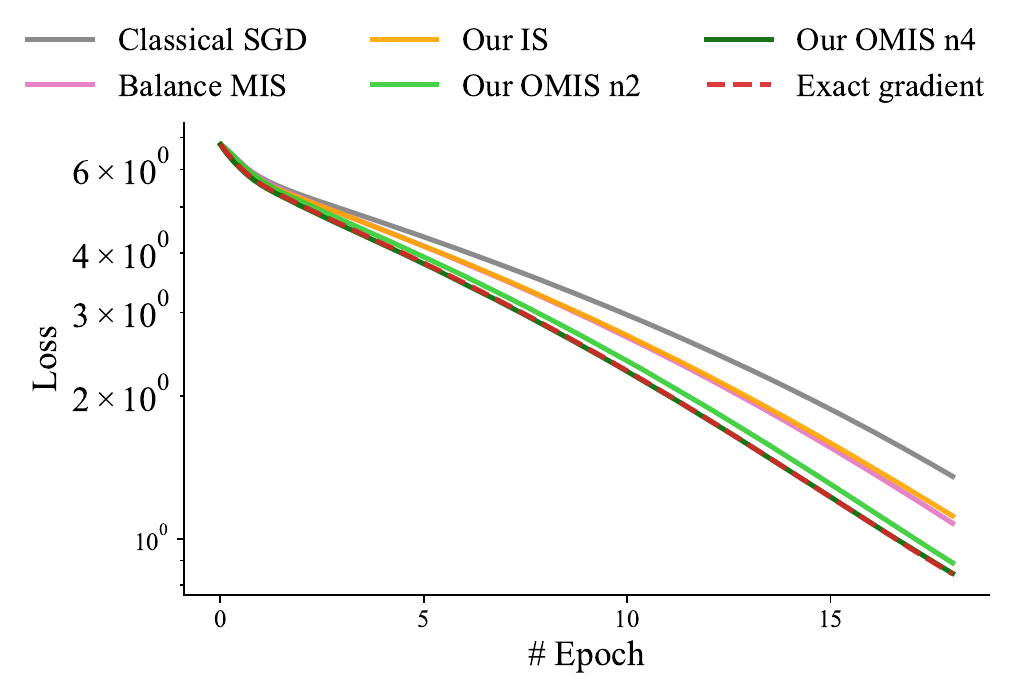}
    }
    \vspace{-2mm}
    \caption{
        Convergence comparison of polynomial regression of order 6 using different method. Exact gradient show a gradient descent as baseline and  classical SGD. For our method, we compare importance sampling and OMIS using $n=2$ or $4$ importance distributions. Balance heuristic MIS is also visible. Our method using OMIS achieve same convergence as exact gradient.
    }
    \label{fig:polynomial_regression}
\end{figure}

\paragraph{Momentum-based linear-system estimation.}

If the matrix estimate $\langle\boldsymbol{A}\rangle$ is inaccurate, its inversion can be unstable and yield a poor gradient estimate. The simplest way to tackle this problem is to use a large number of samples per distribution, which produces a accurate estimates of both $\boldsymbol{A}$ and $\boldsymbol{b}$ and thus a stable solution to the linear system. However, this approach is computationally expensive. Instead, we keep the sample counts low and reuse the estimates from previous mini-batches via momentum-based accumulation, shown in lines 17--18,
where $\beta$ is the parameter controlling the momentum; we use $\beta = 0.7$. This accumulation provides stability, yields an estimate of the momentum gradient \cite{rumelhart1986learning}, and allows us to use 1--4 samples per distribution in a mini-batch.

\paragraph{Importance functions.}

To define our importance distributions, we expand on the approach from \cref{sec:importance_sampling}. Instead of taking the norm of the entire output layer of the model, we take the different gradients separately as $\boldsymbol{\memory}(x) = \frac{\partial \mathcal{L}(x)}{\partial \model(x,\theta)}$ (see \cref{fig:2D_Sampling_visualization}d). Similarly to \cref{alg:IS}, we apply momentum-based accumulation of the per-data importance (line 19 in \cref{alg:OMIS}). If the output layer has more nodes than the desired number $J$ of distributions, we select a subset of the nodes. Many other ways exist to derive the distributions, e.g., clustering the nodes into $J$ groups and taking the norm of each; we leave such exploration for future work.

\begin{figure}[t]
    \centering
    \includegraphics[width=0.9\columnwidth]{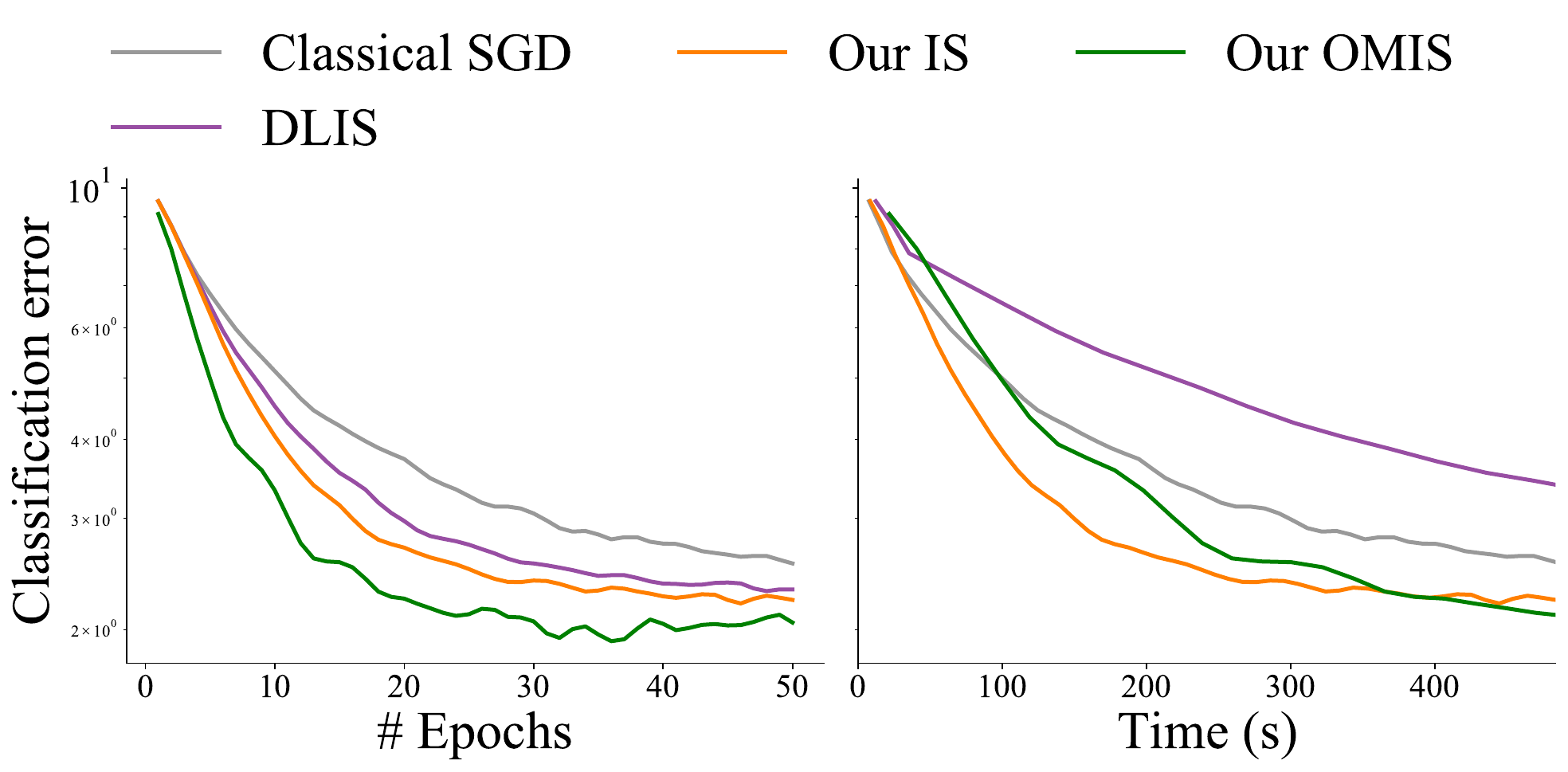}
    \vspace{-2mm}
    \caption{
        Classification error convergence for MNIST classification for various methods. Both \citet{katharopoulos2018dlis}~(DLIS) and resampling SGD approach. In comparison, our two method use the presented algorithm without resampling. It is visible that while DLIS perform similarly to our IS at equal epoch, the overhead of the method makes ours noticeably better at equal time for our IS and OMIS.
    }
    \label{fig:convergence_plots_algo_comp_DLIS}
\end{figure}

\section{Experiments}

\paragraph{Implementation details.}

We evaluate our importance sampling (IS) and optimal multiple importance sampling (OMIS) methods on a set of classification and regression tasks with different data modalities (images, point clouds). We compare them to classical SGD (which draws mini-batch samples uniformly without replacement), DLIS \cite{katharopoulos2018dlis}, and LOW \cite{santiago2021low}. DLIS uses a resampling scheme that samples an initial, larger mini-batch uniformly and then selects a fraction of them for backpropagation and a gradient step. This resampling is based on an importance sampling metric computed by running a forward pass for each initial sample. LOW applies adaptive weighting to uniformly selected mini-batch samples to give importance to data with high loss. All reported metrics are computed on data unseen during training, with the exception of the regression tasks.

All experiments are conducted on a single NVIDIA Tesla A40 graphics card. Details about the optimization setup of each experiment can be found in \cref{app:Optimization_details}.

\begin{figure}[t]
    \centering
    \resizebox{0.9\columnwidth}{!}{
    \begin{tikzpicture}
        \node[inner sep=0pt] (russell) at (6.88,2.9)
        {\includegraphics[scale=0.31]{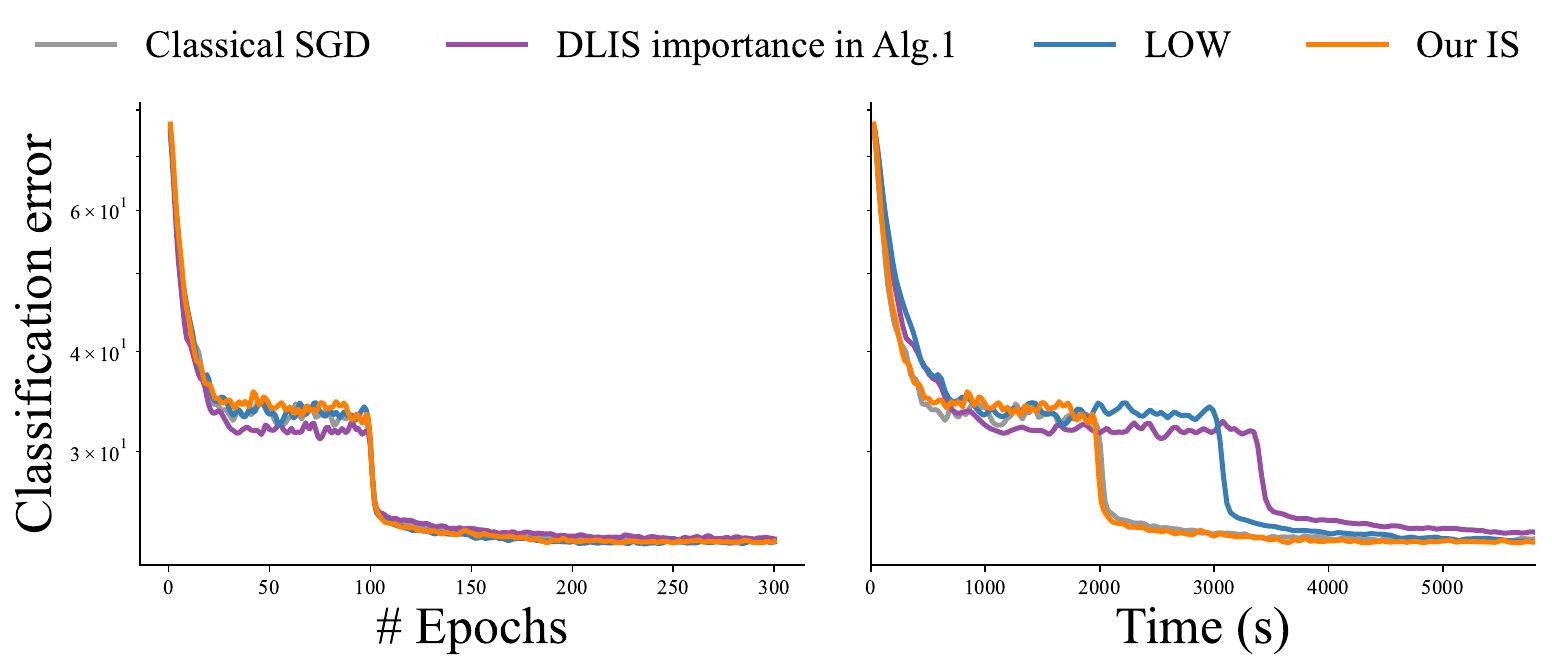}};
        \node[inner sep=0pt] (russell) at (10,3.4)
        {\includegraphics[height=0.9cm]{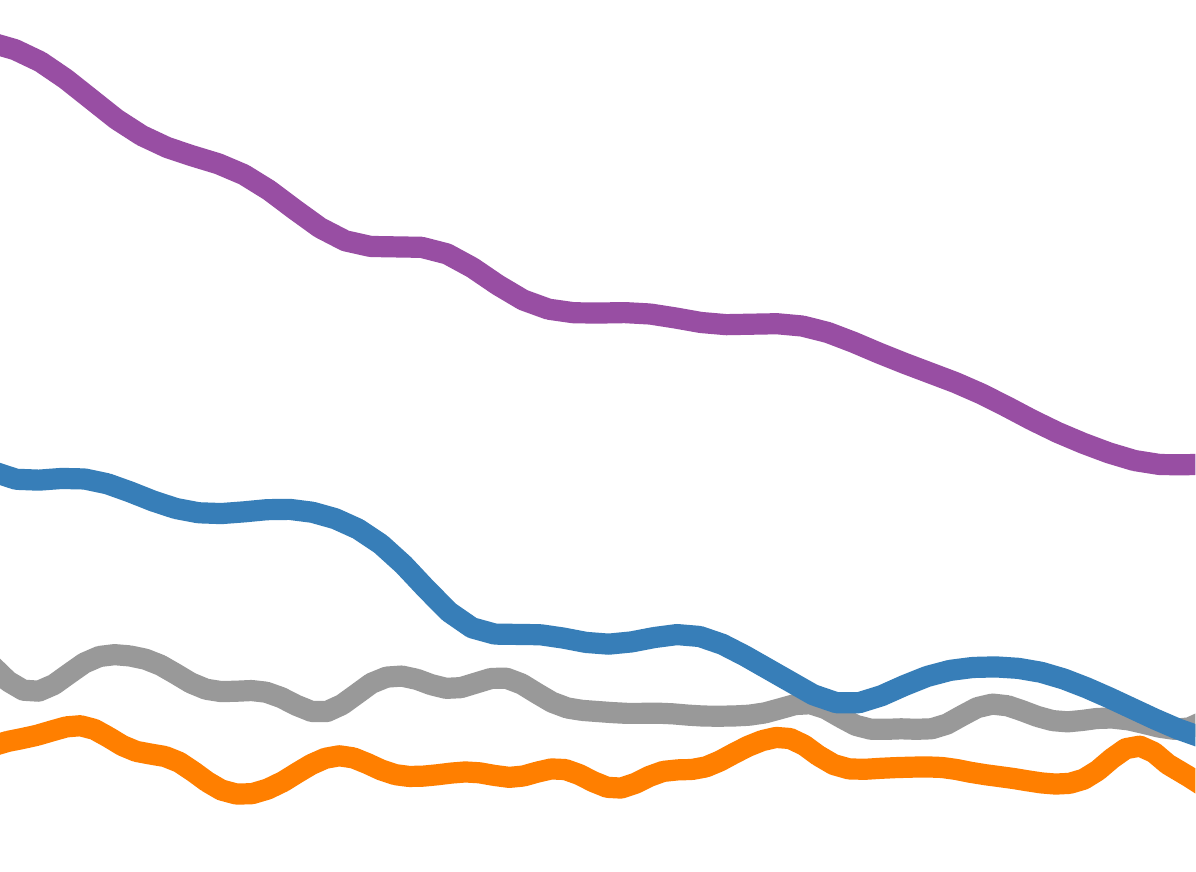}};
        \begin{scope}
            \draw[black,thick] (9.4,2.82) -- (9.4,3.8) -- (10.6,3.8) -- (10.6,2.82) -- cycle;
        \end{scope}
        \draw[black,thick] (9.5,1.75) rectangle (10.8,2.0);
        \draw[black,thick] (10.8,2.0) -- (10.6,2.82);
        \draw[black,thick] (9.5,2.0) -- (9.4,2.82);
        \node[inner sep=0pt] (russell) at (6.,3.3)
        {\includegraphics[height=0.9cm]{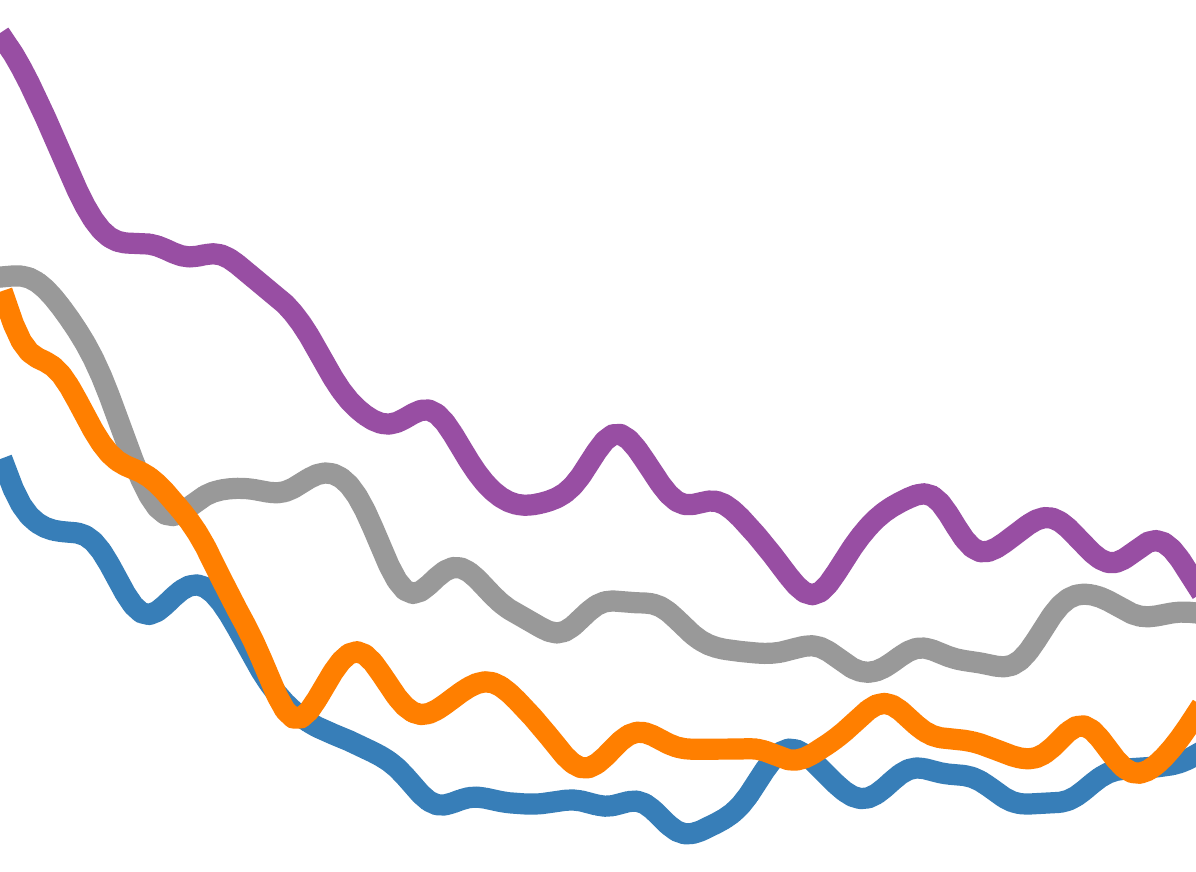}};
        \begin{scope}
            \draw[black,thick] (5.4,2.82) -- (5.4,3.8) -- (6.6,3.8) -- (6.6,2.82) -- cycle;
        \end{scope}
        \draw[black,thick] (5.5,1.75) rectangle (6.8,2.0);
        \draw[black,thick] (6.8,2.0) -- (6.6,2.82);
        \draw[black,thick] (5.5,2.0) -- (5.4,2.82);
    \end{tikzpicture}
    }
    \vspace{-1mm}
    \caption{
        On CIFAR-100, we use the DLIS importance metric in our \cref{alg:IS} instead of the DLIS resampling algorithm. The zoom-in highlights show error drops when the learning rate decreases after epoch 100. Our method (Our IS) outperforms LOW~\cite{santiago2021low} and DLIS weights at equal epochs (left). It also converges faster than LOW and DLIS weights at equal time (right).
    }
    \label{fig:convergence_plots_CIFAR100_DLIS_LOW_equal_time}
\end{figure}

\begin{figure}[t]
    \centering
    \includegraphics[width=0.9\columnwidth]{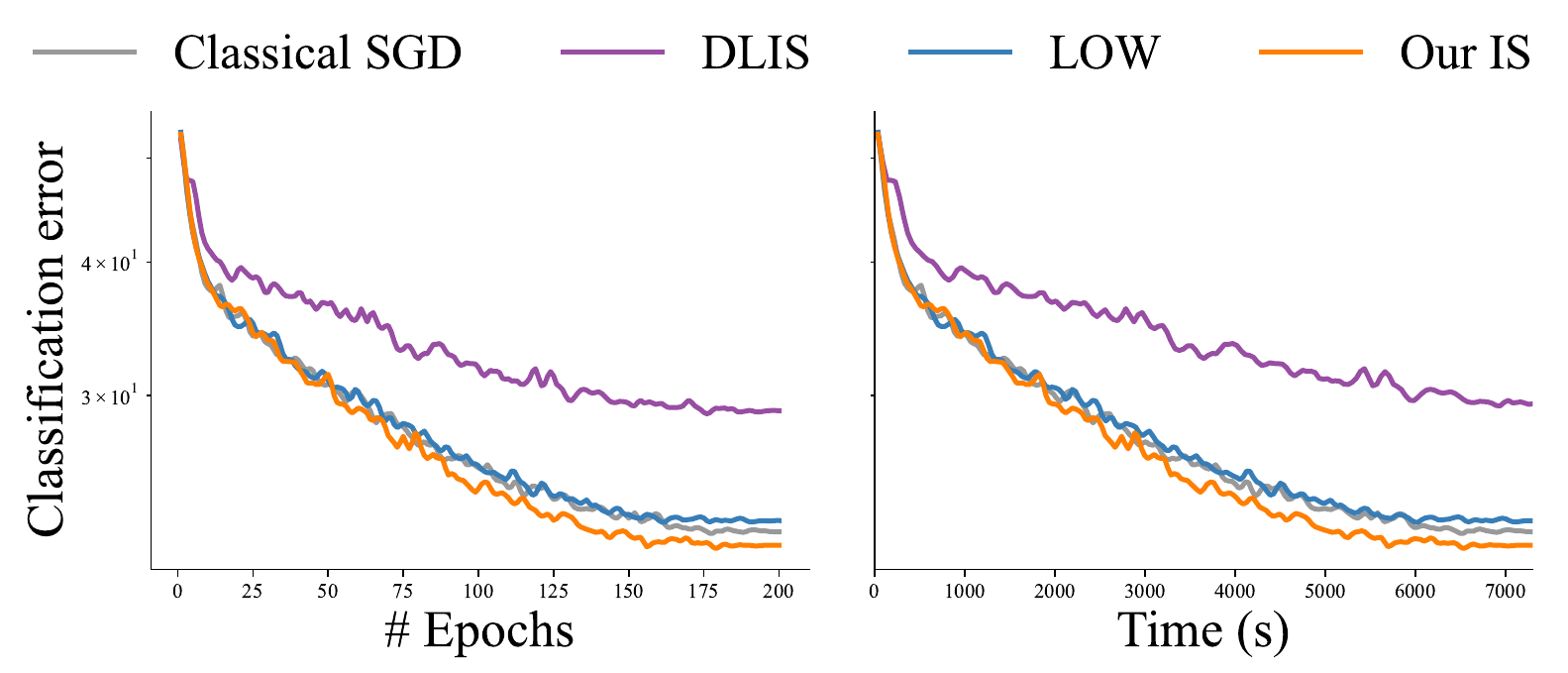}
    \vspace{-2mm}
    \caption{
        Comparisons on CIFAR-10 using Vision Transformer (ViT)~\cite{dosovitskiy2020image}. The results show our importance sampling scheme (Our IS) can improve over classical SGD, LOW~\cite{santiago2021low} and DLIS~\cite{katharopoulos2018dlis} on modern transformer architecture.
    }
    \label{fig:convergence_plots_CIFAR10_tranformer}
    
\end{figure}
\paragraph{Convex problem.}

We performed a basic convergence analysis of IS and OMIS on a convex polynomial-regression problem. \Cref{fig:polynomial_regression} compares classical SGD, our IS, and three MIS techniques: balance heuristic~\cite{veach1997robust} and our OMIS using two and four importance distributions. The exact gradient serves as a reference point for optimal convergence. Balance-heuristic MIS exhibits similar convergence to IS. This can be attributed to the weights depending solely on the relative importance distributions, disregarding differences in individual parameter derivatives. This underscores the unsuitability of the balance heuristic as a weighting method for vector-valued estimation. Both our OMIS variants achieve convergence similar to that of the exact gradient. The four-distribution variant achieves the same quality as the exact gradient using only 32 data samples per mini-batch. This shows the potential of OMIS to achieve low error in gradient estimation even at low mini-batch sizes.

\begin{figure}[t]
    \centering
    \includegraphics[width=0.9\columnwidth]{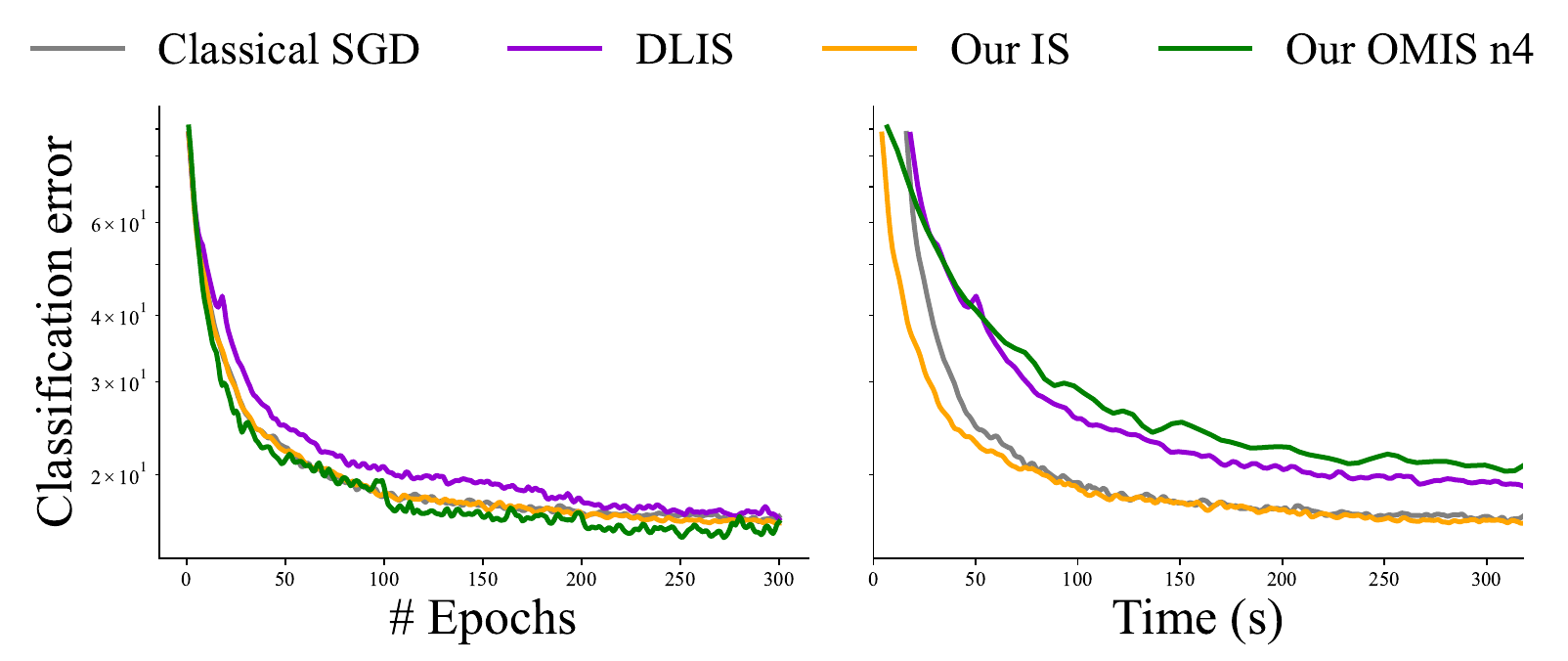}%
    \caption{
        Comparison of our two methods (Our IS, Our OMIS) on point-cloud classification using PointNet~\cite{qi2017pointnet} architecture.
        Our OMIS achieves lower classification error at equal epochs, though it introduces computation overhead as shown at equal-time comparisons. At equal time, our method using importance sampling achieves the best performance.
    }
    \label{fig:Pointcloud_classification_accuracy}
\end{figure}

\begin{figure*}[t]
    \centering
    \resizebox{0.9\textwidth}{!}{
    \newcommand{\PlotTwoImage}[2]{%
        \begin{scope}
            \clip (0,0) -- (1.05,0) -- (1.05,2.5) -- (0,2.5) -- cycle;
           
            \path[fill overzoom image=#1] (0,0) rectangle (2.5cm,2.5cm);
        \end{scope}
        \begin{scope}
            \clip (1.05,0) -- (2.5,0) -- (2.5,2.5) -- (1.05,2.5) -- cycle;
            \path[fill overzoom image=#2] (0,0) rectangle (2.5cm,2.5cm);
        \end{scope}
        \draw (0,0) -- (2.5,0) -- (2.5,2.5) -- (0,2.5) -- cycle;
        \draw (1.05,0) -- (1.05,2.5);
}

\newcommand{\PlotSingleImage}[2]{%

        \begin{scope}
            \clip (0,0) -- (2.5,0) -- (2.5,2.5) -- (0,2.5) -- cycle;
            \path[fill overzoom image=#2] (0,0) rectangle (2.5cm,2.5cm);
        \end{scope}
        \draw (0,0) -- (2.5,0) -- (2.5,2.5) -- (0,2.5) -- cycle;
}

\newcommand{\PlotRef}[1]{%
        \begin{scope}
            \clip (0,0) -- (2.5,0) -- (2.5,2.5) -- (0,2.5) -- cycle;
            \path[fill overzoom image=#1] (0,0) rectangle (2.5cm,2.5cm);
        \end{scope}
        \draw (0,0) -- (2.5,0) -- (2.5,2.5) -- (0,2.5) -- cycle;
        \draw[red]  (1.25/4,1.25/4) -- (3.75/4,1.25/4) -- (3.75/4,3.75/4) -- (1.25/4,3.75/4) -- cycle;
}

\newcommand{\PlotSingleCrop}[1]{%
        \begin{scope}
            \clip (1.25,1.25) -- (3.75,1.25) -- (3.75,3.75) -- (1.25,3.75) -- cycle;
            \path[fill overzoom image=#1] (0,0) rectangle (10cm,10cm);
        \end{scope}
        \draw (1.25,1.25) -- (3.75,1.25) -- (3.75,3.75) -- (1.25,3.75) -- cycle;
}

\newcommand{\PlotSchema}[1]{%
        \begin{scope}
            \path[fill overzoom image=#1] (0,0) rectangle (3.5cm,2.5cm);
        \end{scope}
}

\newcommand{\scaleval}{1.6}
\small
\hspace*{-3mm}
\begin{tabular}{c@{}c@{\;}c@{\;}c@{\;}c@{\;}c@{}}

\multirow{3}{*}[0.74in]{\begin{tikzpicture}[scale=1.2*\scaleval]    
     \PlotSchema{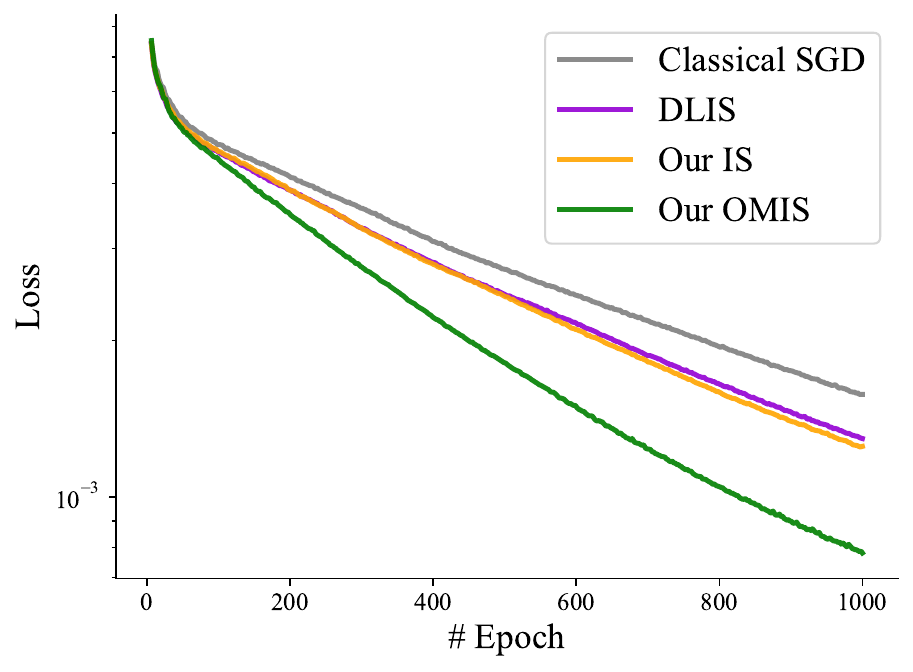}
\end{tikzpicture}}
&
\begin{tikzpicture}[scale=\scaleval/2.05]    
     \PlotRef{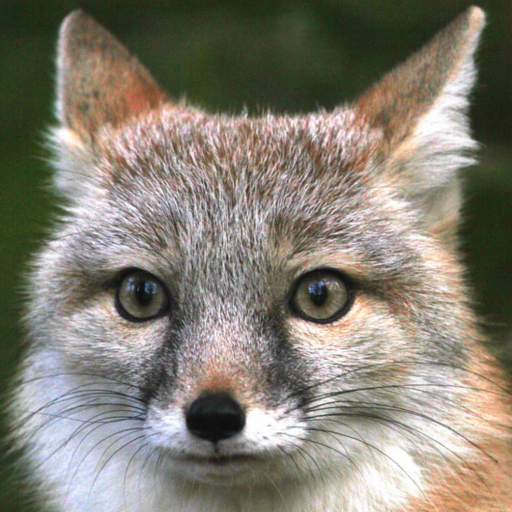}
\end{tikzpicture}
&
\begin{tikzpicture}[scale=\scaleval/2.05]    
     \PlotSingleImage{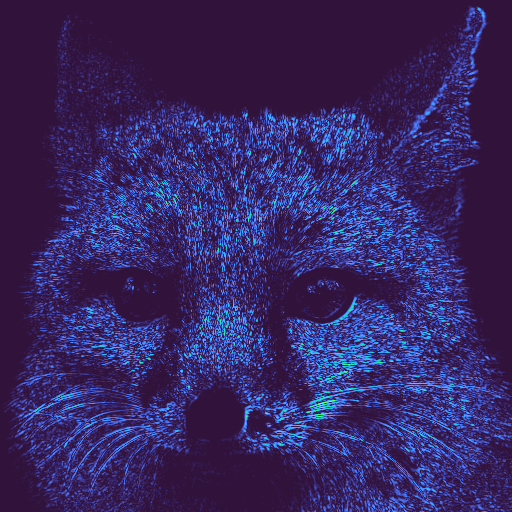}{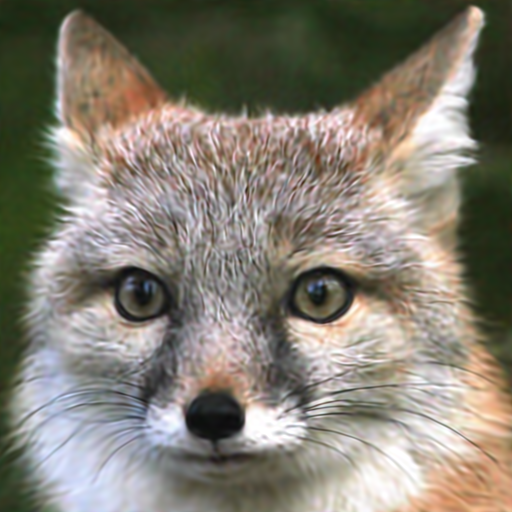}
\end{tikzpicture}
&
\begin{tikzpicture}[scale=\scaleval/2.05]    
     \PlotSingleImage{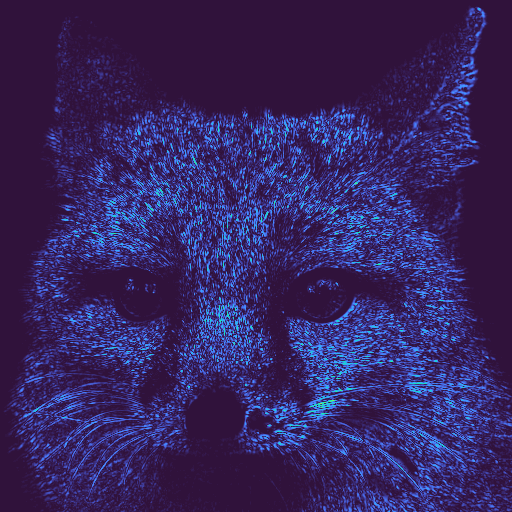}{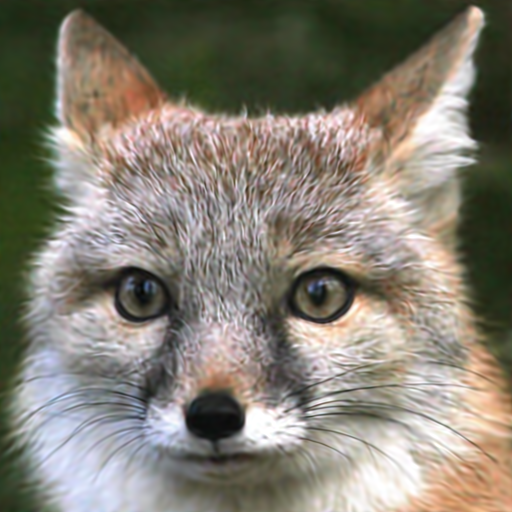}
\end{tikzpicture}
&
\begin{tikzpicture}[scale=\scaleval/2.05]    
     \PlotSingleImage{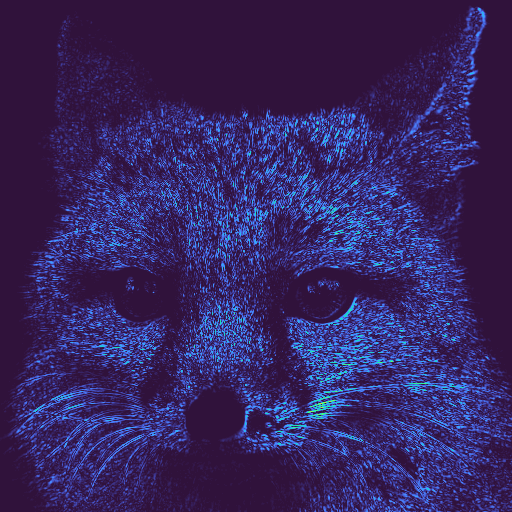}{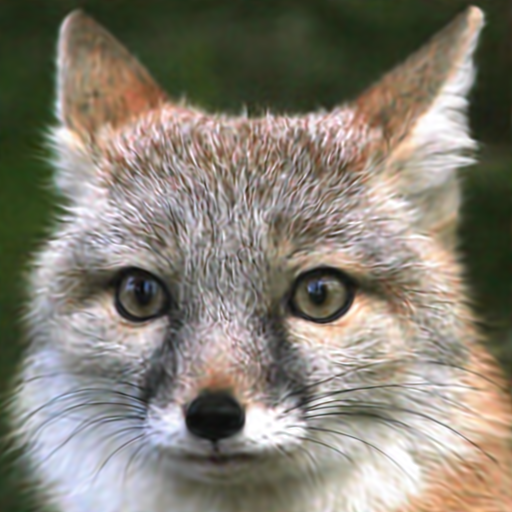}
\end{tikzpicture}
&
\begin{tikzpicture}[scale=\scaleval/2.05]    
     \PlotSingleImage{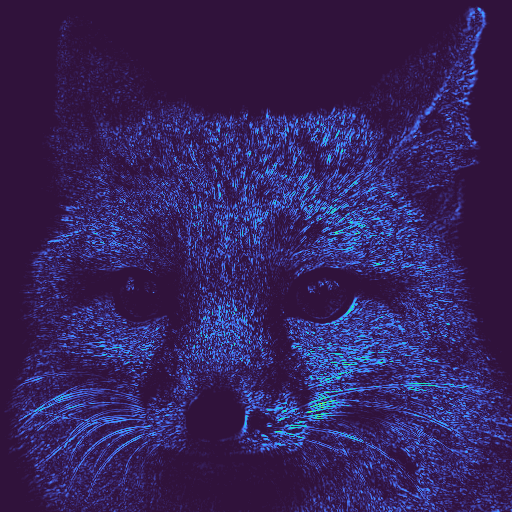}{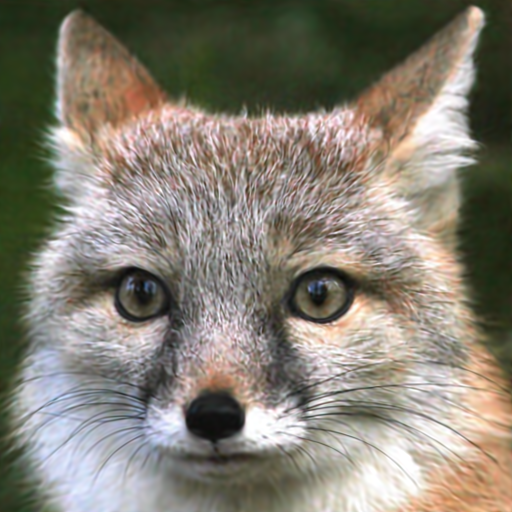}
\end{tikzpicture}
\\
~
&
\begin{tikzpicture}[scale=\scaleval/2.05]    
     \PlotSingleCrop{figures/2D_nerf/ref.png}
\end{tikzpicture}
&
\begin{tikzpicture}[scale=\scaleval/2.05]    
     \PlotSingleCrop{figures/2D_nerf/Default_n1.png}
\end{tikzpicture}
&
\begin{tikzpicture}[scale=\scaleval/2.05]    
     \PlotSingleCrop{figures/2D_nerf/DLIS_n1.png}
\end{tikzpicture}
&
\begin{tikzpicture}[scale=\scaleval/2.05]    
     \PlotSingleCrop{figures/2D_nerf/OptiMIS_n1.png}
\end{tikzpicture}
&
\begin{tikzpicture}[scale=\scaleval/2.05]    
     \PlotSingleCrop{figures/2D_nerf/OptiMIS_n3.png}
\end{tikzpicture}
\\
~ &Reference & Uniform & DLIS & \textbf{Our IS} & \textbf{Our OMIS} \\
\end{tabular}
    }
    \vspace{1mm}
    \caption{
        Comparison at equal step for image 2D regression. Left side show the convergence plot while the right display the result regression and a close-up view. Our method using MIS achieves the lower error on this problem while IS and DLIS perform similarly. On the images it is visible that our OMIS recover the finest details of the fur and whiskers.
    }
    
    \label{fig:2D_nerf}
\end{figure*}

\paragraph{Classification.}

In~\cref{fig:convergence_plots_algo_comp_DLIS}, we compare our algorithms to the DLIS resampling algorithm of \citet{katharopoulos2018dlis} on MNIST classification. Our IS performs slightly better than DLIS, and our OMIS does best. The differences between our methods and the rest are more pronounced when comparing equal-time performance. DLIS has a higher computational cost as it involves running a forward pass on a large mini-batch to compute resampling probabilities. Our OMIS requires access to the gradient of each mini-batch sample; obtaining these gradients in our current implementation is inefficient due to technical limitations in the optimization framework we use (PyTorch). Nevertheless, the method manages to make up for this overhead with a higher-quality gradient estimate. In~\cref{fig:convergence_plots_algo_comp_DLIS} we compare classification error; loss-convergence plots are shown in \cref{app:extra_results} (\cref{fig:convergence_plots_algo_comp_DLIS_MINIS_loss}).

In~\cref{fig:convergence_plots_CIFAR100_DLIS_LOW_equal_time}, we compare our IS against using the DLIS importance function in \cref{alg:IS} and LOW~\cite{santiago2021low} on CIFAR-100 classification. At equal number of epochs, the difference between the methods is small (see close-up view). Our IS achieves similar classification accuracy as LOW and outperforms the DLIS variant. At equal time the difference is more important as our method has lower computational cost. This experiment shows that our importance function achieves better performance than that of DLIS within the same optimization algorithm. 

\Cref{fig:convergence_plots_CIFAR10_tranformer} shows a similar experiment on CIFAR-10 using a vision transformer~\cite{dosovitskiy2020image}. Our IS method achieves consistent improvement over the state of the art. The worse convergence of (original, resampling-based) DLIS can be attributed to its resampling tending to exclude some training data with very low importance, which can cause overfitting.

\Cref{fig:Pointcloud_classification_accuracy} shows point-cloud classification, where our IS is comparable to classical SGD and our OMIS outperforms other methods in terms of classification error at equal epochs. In complex cases where importance sampling cannot enhance convergence by providing a more accurate gradient estimator, our method is still as efficient as SGD due to minimal overhead. This means that even though importance sampling does not offer additional benefits in these scenarios, our implementation remains competitive with classical methods. In his case DLIS and our OMIS both suffer from computational overhead.%

We also perform an ablation study for linear-system momentum in~\cref{alg:OMIS}. We apply same momentum on the gradient for classical SGD, DLIS and our IS. \Cref{app:extra_results} (\cref{fig:Pointnet_classification_lossAccuracy}) shows this comparison. Our OMIS still outperforms other methods for this task at equal steps.

\paragraph{Regression.}

\Cref{fig:2D_nerf} shows results on image regression, comparing classical SGD, DLIS, and our IS and OMIS. Classical SGD yields a blurry image, as seen in the zoom-ins. DLIS and our IS methods achieves similar results, with increased whisker sharpness but still blurry fur, though ours has slightly lower loss and is computationally faster, as discussed above. Our OMIS employs three sampling distributions based on the network's outputs which represent the red, green and blue image channels. This method achieves the lowest error and highest image fidelity, as seen in the zoom-in.

\section{Limitations and future work}

We have showcased the effectiveness of importance sampling and optimal multiple importance sampling (OMIS) in machine-learning optimization, leading to a reduction in gradient-estimation error. Our current OMIS implementation incurs some overhead as it requires access to individual mini-batch sample gradients. Modern optimization frameworks can efficiently compute those gradients in parallel but only return their average. This is the main computational bottleneck in the method. The overhead of the linear system computation is negligible; we have tested using up to 10 distributions.

Our current OMIS implementation is limited to sequential models; hence its absence from our ViT experiment in \cref{fig:convergence_plots_CIFAR10_tranformer}. However, there is no inherent limitation that would prevent its use with such more complex architectures. We anticipate that similar improvements could be achieved, but defer the exploration of this extension to future work. 

In all our experiments we allocate the same sampling budget to each distribution. Non-uniform sample distribution could potentially further reduce estimation variance, especially if it can be dynamically adjusted during the optimization process.

Recent work from \citet{santiago2021low} has explored a variant of importance sampling that forgoes sample-contribution normalization, i.e., the division by the probability $p(x)$ in \cref{eq:MC_Gradient} (and on line 10 of \cref{alg:IS}). This heuristic approach lacks proof of convergence but can achieve practical improvement over importance sampling in some cases. We include a such variant of our IS method in \cref{app:extra_results}.

\section{Conclusion}

This work proposes a novel approach to improve gradient-descent optimization through efficient data importance sampling. We present a method incorporates a gradient-based importance metric that evolves during training. It boasts minimal computational overhead while effectively exploiting the gradient of the network output. Furthermore, we introduce the use of (optimal) multiple importance sampling for vector-valued, gradient estimation. Empirical evaluation on typical machine learning tasks demonstrates the tangible benefits of combining several importance distributions in achieving faster convergence.

\bibliographystyle{apalike}
{\small
\bibliography{example}}

\appendix

\section{Optimization details}
\label{app:Optimization_details}

\paragraph{Classification.}

The classification tasks include image classification (MNIST~\cite{deng2012mnist}, CIFAR-10/100~\cite{krizhevsky2009learning} and point-cloud (ModelNet40~\cite{wu20153d}) classification. 

The MNIST database contains 60,000 training images and 10,000 testing images. 
We train a 3-layer fully-connected network (MLP) for MNIST over 50 epochs with an Adam optimizer \cite{kingma2014adam}. CIFAR-10, introduced by \citet{krizhevsky2009learning}, is a dataset that consists of 60,000 color images of size 32x32. These images belong to 10 different object classes, each class having 6,000 images. On the other hand, CIFAR-100~\citet{krizhevsky2009learning} contains 100 classes with 600 images each. For each class, there are 500 training images and 100 testing images. In our experiments, we train the ResNet-18 network \cite{he2016deep} on both datasets. 
We apply random horizontal flip and random crops to augment the data during training.
ModelNet40 contains 9,843 point clouds for training and 2,468 for testing.
Each point cloud has 1,024 points. 
We train a PointNet~\cite{qi2017pointnet} with 3 shared MLP layers and 2 fully-connected layers for 300 epochs on point-cloud classification.
We use the Adam optimizer~\citet{kingma2014adam}, with batch size 64, weight decay 0.001, initial learning rate 0.00002 divided by 2 after 100, 200 epochs.

\paragraph{Regression.}

Polynomial regression consists of optimizing the coefficients of a 1D polynomial of a given order to fit randomly drawn data from a reference polynomial of the same order. The reference data are generated on the interval $[-2;2]$. Optimization is done using an Adam optimizer~\cite{kingma2014adam} with a mini-batch size of 32 elements.

The image regression task consists in learning the mapping between a 2D coordinate input (pixel coordinate) and the 3-color output of the image for this pixel. We use a network with 5 fully-connected layers associated with positional encodings using SIREN activations~\cite{sitzmann2020implicit}. The training is done over 500 epoch using an Adam~\cite{kingma2014adam} optimizer and each mini-batch is composed of $256$ pixels for a $512^2$ reference image.

\section{Multiple importance sampling in brief}
\label{app:MIS}

\paragraph{Importance sampling.}

An Importance sampling Monte Carlo estimator $\langle F \rangle_\mathrm{IS}$ of a function $f$ is define as :
\begin{equation}
    \langle F \rangle_\mathrm{IS} = \sum_{i=1}^{n}\frac{f(x_{i})}{n p(x_{i})},\qquad x_i \propto \pdf(x).
\end{equation}\\
With $x_i$ the $i^{th}$ data sample drawn following the probability distribution function $\pdf(x)$.

The effectiveness of this estimator depends on the relation between the functions $f(x)$ and $\pdf(x)$. The variance of such estimator is :
\begin{equation}
    \Var{[\langle F \rangle_\mathrm{IS}]} = \frac{1}{n}\Var{[\nicefrac{f}{\pdf}]}.
    \label{eq:Var_IS_Monte_Carlo_estimator}
\end{equation}\\
Reducing variance in the estimation depends on the proportionality between the function $f$ and the probability density $\pdf$. 

When dealing with multivariate functions, finding a probability density proportional to every parameters is often impractical. A trade-off is required to obtain a single probability distribution maximizing the proportionality with all the parameters of the function simultaneously. Several studies, such as \cite{zhaoa2015stochastic,needell2014stochastic,wang2017accelerating,alain2015variance}, have shown that the optimal choice of sampling strategy is the $L_2$ norm of the function $f$.

\paragraph{Multiple importance sampling.}

Multiple Importance Sampling (MIS) is a technique that combines multiple sampling strategies with associated weightings, unlike importance sampling which relies on a single strategy. This approach allows for a more versatile gradient estimation. The MIS Monte Carlo estimator, denoted as $\langle F \rangle_\mathrm{MIS}$, is calculated by summing over all samples drawn independently for each strategy, and then using a weighted estimator. The equation for $\langle F \rangle_\mathrm{MIS}$ is given by:

\begin{equation}
    \langle F \rangle_\mathrm{MIS} = \sum_{j=1}^{J}\sum_{i=1}^{n_j}w_{j}(x_{ij})\frac{f(x_{ij})}{n_jp_j(x_{ij})}
    \label{eq:MIS_Monte_Carlo_estimator_appendix}
\end{equation} 

Here, $x_{ij}$ represents the $i^{th}$ sample from the $j^{th}$ technique, $w_j(x)$ is a weighting function such that $f(x) \neq 0 \Rightarrow \sum^J_{j=1}w_j(x) = 1$, and $p_j(x)=0 \Rightarrow w_j(x) = 0$. $J$ is the number of sampling techniques, and $n_j$ is the number of samples generated by the $j^{th}$ technique. The variance of a Monte Carlo estimator using MIS, denoted as $\Var[\langle F \rangle_\mathrm{MIS}]$, can be expressed as: 

\begin{equation}
    \Var[\langle F \rangle_\mathrm{MIS}] = \sum_{j=1}^{J} \int_D \frac{w_j(x)^2f(x)^2}{n_jp_j(x)}dx - \sum_{j=1}^{J} \frac{1}{n_j} \langle w_j, f \rangle ^2
    \label{eq:MISvariance}
\end{equation} 

The balance heuristic~\cite{veach1997robust} is the most commonly used MIS heuristic. It sets the weight of the samples from each technique according to the following equation: 

\begin{equation}
    w_{j}(x_{i}) = \frac{n_jp_j(x_i)}{\sum_{k=1}^{J}n_kp_k(x_i)}
    \label{eq:BalanceW}
\end{equation} 

This weighting strategy effectively mitigates the impact of events with low probability when samples are drawn from a low-probability distribution. It prevents a large increase in the contribution of such events in the Monte Carlo estimator \eqref{eq:MIS_Monte_Carlo_estimator_appendix} where the function value would be divided by a very low value. The balance heuristic compensates for this and avoids extreme cases. Overall, this weighting strategy increases the robustness of the importance sampling estimator, but it is limited by its independence from the function value.

\paragraph{Optimal weighting.}

Following the discussion in~\cref{sec:optimal_mis}, 
it can also be deduced from \cref{eq:MIS_Monte_Carlo_estimator_appendix,eq:OptiMISweigth} that $\langle F \rangle_\mathrm{OMIS} = \sum_{j=1}^J \alpha_j$. Given a set of probability distribution functions $p_1$, \ldots, $p_J$, we can formulate the optimal MIS solver as \cref{alg:OptiMISpseudoCode}. $\boldsymbol{W}_{ij}$ represents the vector containing the balance weight (\ref{eq:BalanceW}) w.r.t.\ the J sampling techniques and the normalization factor $S(x_{ij}) = \sum_{k = 1}^J n_k \pdf_k(x_{ij})$.

\begin{algorithm}[t]
    \caption{
        Optimal multiple importance sampling solver.
    }
    \label{alg:OptiMISpseudoCode}
    \begin{algorithmic}[1]
        \State $\langle\boldsymbol{A} \rangle \leftarrow 0^{J \times J} , \langle\boldsymbol{b} \rangle \leftarrow 0^{J}$
        \Forloop{$t \leftarrow 1$ to $T$}
        \Forloop{$j \leftarrow 1$ to $J$}
        \State $ \{ x_{ij}\}_{i=1}^{n_j} \leftarrow $ draw $n_j$ samples from technique $p_j$
        \EndForloop
        \State $\langle\boldsymbol{A}\rangle \leftarrow \langle\boldsymbol{A}\rangle  + \sum_{j = 1}^J \sum_{i = 1}^{n_j} \boldsymbol{W}_{ij}\boldsymbol{W}_{ij}^T$
        \State $\langle \boldsymbol{b} \rangle \leftarrow \langle\boldsymbol{b} \rangle  + \sum_{j = 1}^J \sum_{i = 1}^{n_j} f(x_{ij})\boldsymbol{W}_{ij}/S(x_{ij})$
        \EndForloop
        \State $\langle \boldsymbol{\alpha} \rangle \leftarrow \text{solve linear system  } \langle \boldsymbol{A} \rangle \langle \boldsymbol{\alpha} \rangle = \langle \boldsymbol{b} \rangle$
        \State \Return $\sum_{j = 1}^N \langle \boldsymbol{\alpha_j} \rangle$
    \end{algorithmic}
\end{algorithm}

The algorithm proceeds through three key stages. The first stage involves initializing the linear system defined in \cref{eq:AlphaDefinition} (line 1). The second stage iteratively updates the system for each drawn data sample (lines 5-6). Upon completion of this process, the matrix $\boldsymbol{A}$ and vector $\boldsymbol{b}$ provide Monte Carlo approximations of the quantities specified in \cref{eq:AlphaDefinition}. The third and final stage involve solving the linear system to obtain the vector $\boldsymbol{\alpha}$ (line 7). The estimated value of $\langle F \rangle_\mathrm{MIS}^o$ is then returned.

It can be noted that the linear system size scales with the number of sampling techniques. More importantly each sampling technique needs to be sampled in order create a linear system possible to solve. The number a sample of each technique does not have to be the same but requires to be fixed at the start of the algorithm.
Also the presented algorithm works for a scalar value function. In the case of multivariate function, multiple contribution vector $\boldsymbol{b}$ need to be constructed (one per parameter) and the linear system needs to be solved for each.

\section{Algorithm details}
\label{app:is_algorithm_details}
 This section presents the two initialization subroutine for~\cref{alg:IS} and~\cref{alg:OMIS}. The role of the methods is to run a first epoch in a classical SGD loop in order to process every data once. For each data the importance metric is reported into the memory $\memory$ and returned with the current model parameters. This approach avoids computing the importance for all data without benefiting from the required forward step computed. 
 
\begin{algorithm}[h]
    \caption{SGD-based initialization of persistent per-data importance $\memory$ in \cref{alg:IS}.}
    \label{alg:Initialization}
    \begin{algorithmic}[1]
        \Function{Initialize}{$x$,$y$,$\Omega$,$\theta$,$B$} 
            \Forloop{$t \leftarrow 1$ \textbf{to} $|\Omega|/B$} %
                \State $x,y \leftarrow \{x_i, y_i \}_{i=(t-1)\cdot\batchSize+1}^{t
                \cdot\batchSize+1}$ 
                \State $l(x) \leftarrow \loss(m(x,\theta),y)$
                \State $\nabla l(x) \leftarrow $ Backpropagate$(l(x))$ 
                \State $\totallossEst{\theta}(x) \leftarrow $ $\nabla l(x)/B$ \algComment{\cref{eq:MC_Gradient}}
                \State $\theta \leftarrow \theta - \eta \, \totallossEst{\theta}(x)$ \algComment{\cref{eq:gradientDescent}}
                \State $\memory(x) \leftarrow \left\Vert \frac{\partial \mathcal{L}(x)}{\partial \model(x,\theta)} \right\Vert$
            \EndForloop
            \State \Return $q$,$\theta$
        \EndFunction   
    \end{algorithmic}
\end{algorithm}

\begin{algorithm}[h]
    \caption{Subroutine for initialization in \cref{alg:OMIS}}
    \label{alg:Initialization_MIS}
    \begin{algorithmic}[1]
        \Function{InitializeMIS}{$x$,$y$,$\Omega$,$\theta$,$B$} 
        \State \algCommentNoArrow{Initialize $\boldsymbol{\memory}$ in a classical SGD loop}
            \Forloop{$t \leftarrow 1$ \textbf{to} $|\Omega|/B$} %
                \State $x,y \leftarrow \{x_i, y_i \}_{i=(t-1)\cdot\batchSize+1}^{t
                \cdot\batchSize+1}$ 
                \State \algCommentNoArrow{See all samples in the first epoch}
                \State $l(x) \leftarrow \loss(m(x,\theta),y)$
                \State $\nabla l(x) \leftarrow $ Backpropagate$(l(x))$ 
                \State $\totallossEst{\theta}(x) \leftarrow $ $\nabla l(x)/B$ \algComment{\cref{eq:MC_Gradient}}
                \State $\theta \leftarrow \theta - \eta \, \totallossEst{\theta}(x)$ \algComment{\cref{eq:gradientDescent}}
                \State $\boldsymbol{\memory}(x) \leftarrow  \frac{\partial \mathcal{L}(x)}{\partial \model(x,\theta)} $
            \EndForloop
            \State \Return $\boldsymbol{\memory}$,$\theta$
        \EndFunction   
    \end{algorithmic}
\end{algorithm}

\begin{algorithm}[h]
    \small
    \caption{Subroutine for cross entropy loss importance metric}
    \label{alg:ComputeSampleImportance}
    \begin{algorithmic}[1]
        \State {$x_i =$ data sample, $y_i =$ class index of $x_i$}
        \Function{Importance}{$x_i$,$y_i$} 
            \State $s \leftarrow \exp(\outputlayer)/\sum_{k=1}^{\classcount} \exp(\outputlayer_k)$  \algComment{Eq.\ref{eq:CrossEntropyLoss}}
            \State $q \leftarrow \sum_{j=1}^\classcount s_j - \mathbf{1}_{j = y_i}$ \algComment{Eq.\ref{eq:CrossEntropyGrad}}
            \State \Return $q$
        \EndFunction
    \end{algorithmic}
\end{algorithm}

\section{Cross-entropy loss gradient}
\label{app:Cross_entropy_close_form}

Machine learning frameworks take data $x$ as input, perform matrix multiplication with weights and biases added. 
The output layer is then fed to the softmax function to obtain values $\score$ that are fed to the loss function. 
$\target$ represents the target values.
We focus on the categorical cross-entropy loss function for the classification problem (with $\classcount$ categories) given by:
\begin{multline}
    \label{eq:CrossEntropyLoss}
    \lossce = - \sum_i \target_i \log \score_i,
    \;\;\;
    \score_i = \frac{\exp(\outputlayer_l)}{\sum_l^\classcount \exp(\outputlayer_l)}.\!
\end{multline}
For backpropagation, we need to calculate the derivative of the $\log\score$ term w.r.t.\ the weighted input $z$ of the output layer. We can easily derive the derivative of the loss from first principles as shown below:
\begin{equation}
\begin{split}
    \frac{\partial \lossce}{\partial \outputlayer_j}
    &= - \frac{\partial}{\partial \outputlayer_j} \left( \sum_i^{\classcount} \target_i \log \score_i \right)\\
    &= - \sum_i^{\classcount} \target_i \frac{\partial}{\partial \outputlayer_j} \log \score_i \\
    &= - \sum_i^{\classcount} \frac{\target_i}{\score_i} \frac{\partial \score_i}{\partial \outputlayer_j} \\
    &= - \sum_i^{\classcount} \frac{\target_i}{\score_i} \score_i \cdot (\unitfunc\{ i == j \} - \score_j) \\
    &= \sum_i^{\classcount} {\target_i} \cdot \score_j - \sum_i^{\classcount} \target_i \cdot (\unitfunc\{ i == j \}) \\
    &= \score_j \sum_i^{\classcount} {\target_i}  -  \target_j = \score_j - \target_j
\end{split}
\end{equation}
The partial derivative of the cross-entropy loss function w.r.t.\ output layer parameters has the form:
\begin{align}
\label{eq:CrossEntropyGrad}
        \frac{\partial \lossce}{\partial \outputlayer_j} &= \score_j - \target_j
\end{align}
For classification tasks, we directly use this analytic form of the derivative and compute it's norm as weights for adaptive and importance sampling.

\section{Importance momentum}
\label{sec:importance_momentum}

Updating the persistent per-sample importance $\memory$ directly sometime leads to a sudden decrease of accuracy during training. 
To make the training process more stable, we update $\memory$ by linearly interpolating the importance at the previous and current steps:
\begin{equation}
    \memory(x) = \momentum \cdot \memory_{prev}(x) + (1 - \momentum) \cdot \memory(x)
    \label{eq:Weight_momentum}
\end{equation}
where $\momentum$ is a constant for all data samples. In practice, we use  $\momentum \in \{0.0, 0.1, 0.2, 0.3\}$ as it gives the best trade-off between importance update and stability. This can be seen as a momentum evolution of the per-sample importance to avoid high variation. 
{Utilizing an exponential moving average to update the importance metric prevents the incorporation of outlier values. This is particularly beneficial in noisy setups, like situations with a high number of class or a low total number of data.}

\section{Additional results}
\label{app:extra_results}

\begin{figure}[h]
    \centering
    \hfill
    \resizebox{\columnwidth}{!}{
    {\includegraphics[width=0.495\textwidth]%
{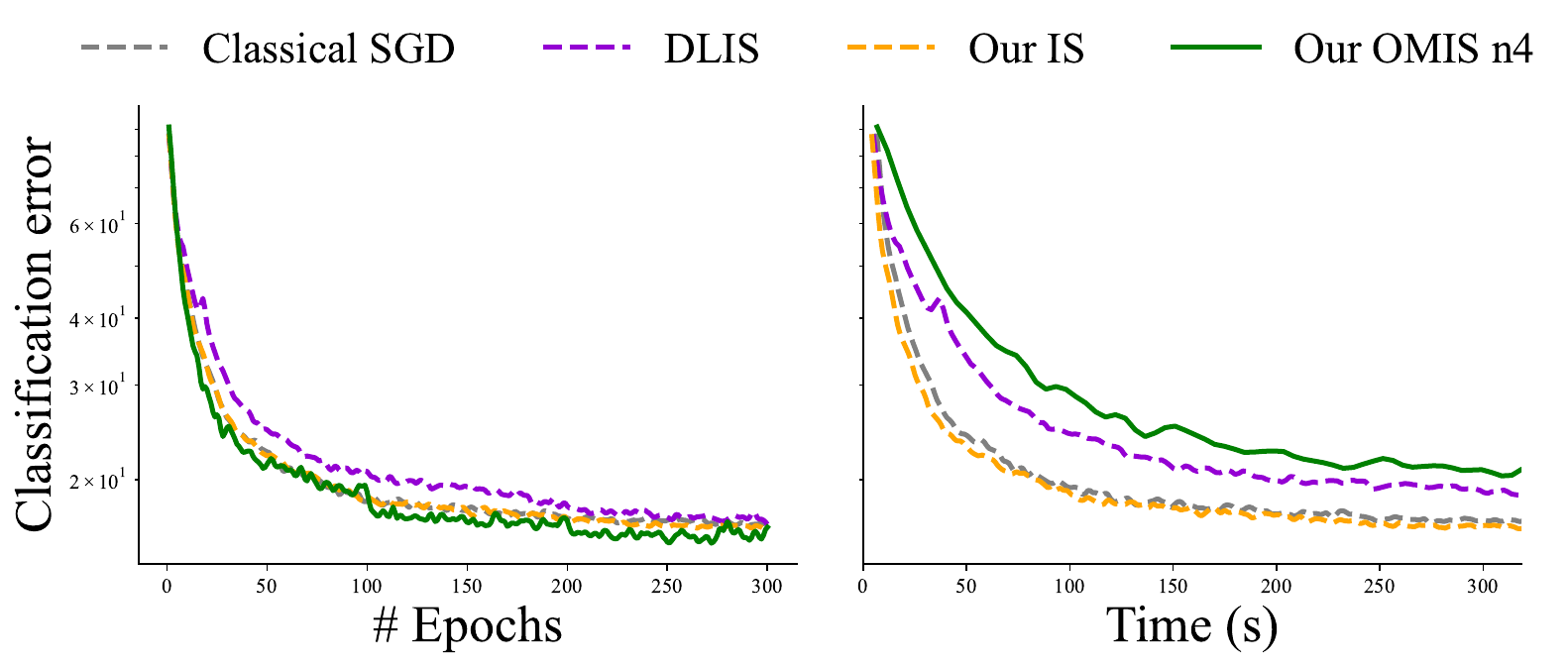}}}
    \hfill
    \caption{
        Ablation study on point-cloud classification using linear-system momentum as described in~\cref{alg:OMIS} for baselines represented as dashed lines. Our OMIS still outperforms other baselines at equal epochs, similar to the results shown in~\cref{fig:Pointcloud_classification_accuracy}.
    }
    \label{fig:Pointnet_classification_lossAccuracy}
\end{figure}

This section provides additional results, including an ablation study as shown in~\cref{fig:Pointnet_classification_lossAccuracy} for linear-system momentum used in~\cref{alg:OMIS} and results of our adaptive sampling method.
~\cref{fig:Pointnet_classification_lossAccuracy,fig:Pointcloud_classification_accuracy} demonstrate that classical SGD, DLIS and Our IS work similarly with and without momentum. Our OMIS outperforms other methods in both cases.

\begin{figure}[h]
    \centering
    \resizebox{\columnwidth}{!}{
    \includegraphics[scale=0.248]{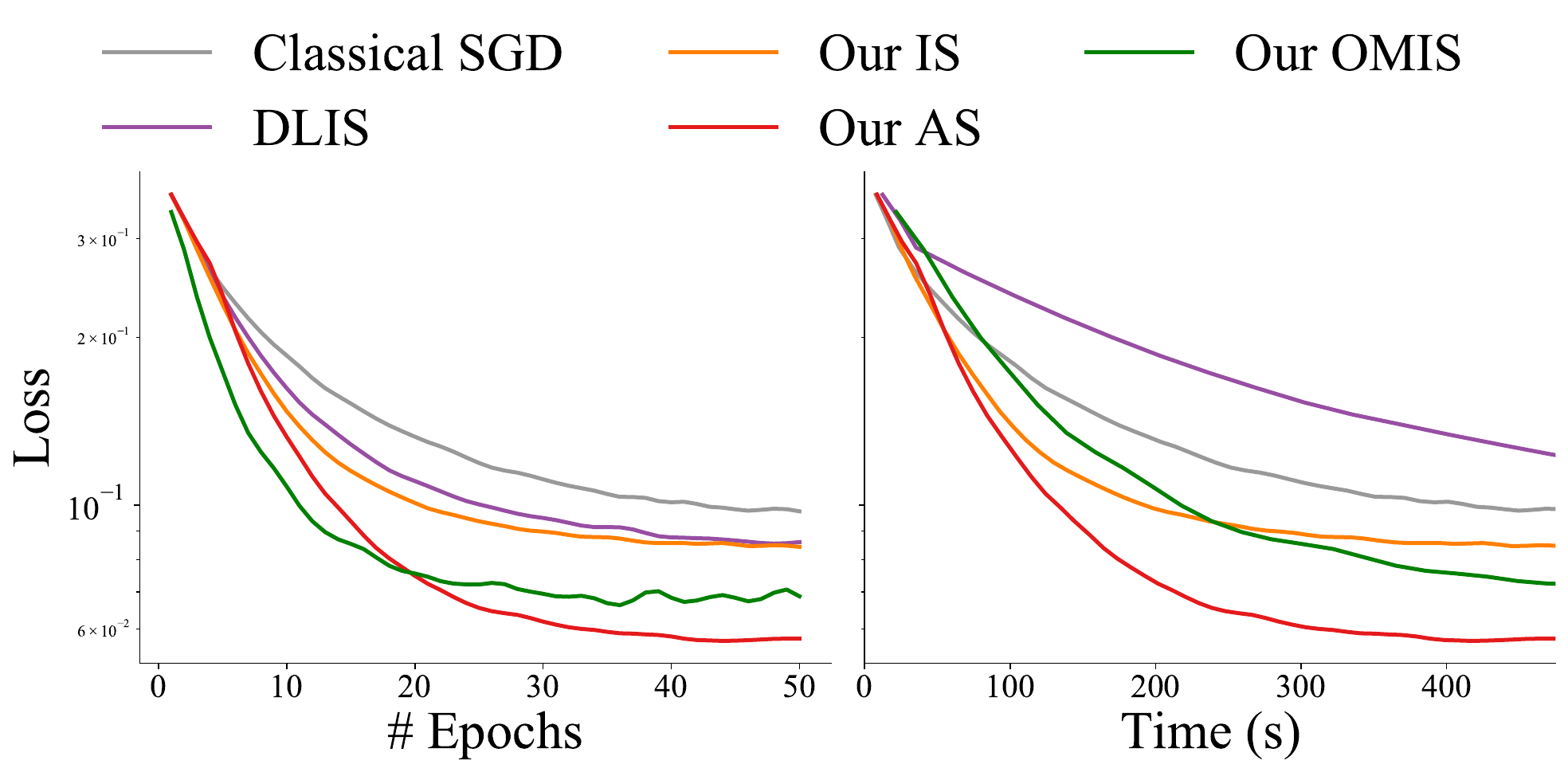}}
    \caption{We compare loss for the MNIST dataset between the resampling algorithm by  \citet{katharopoulos2018dlis} (DLIS) and our algorithm. 
    At equal epochs, DLIS works better than both classical and resampling SGD. 
    However, at equal time, the resampling cost is too high, making DLIS even slower than standard SGD.
    }
    \label{fig:convergence_plots_algo_comp_DLIS_MINIS_loss}
\end{figure}

\begin{figure}[h]
    \centering
    \resizebox{\columnwidth}{!}{
    \includegraphics[scale=0.32]%
    {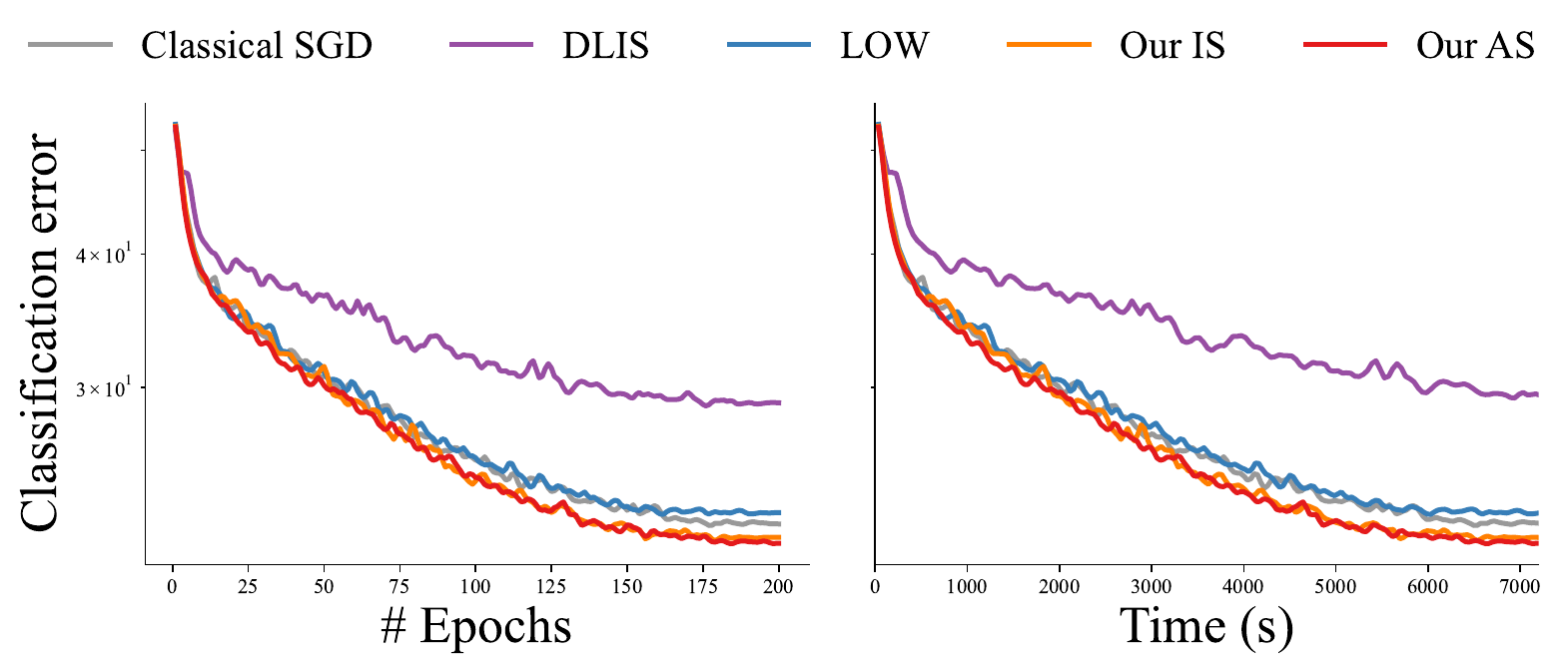}}
    \caption{Comparisons on CIFAR-10 using Vision Transformer (ViT)~\cite{dosovitskiy2020image}. The results show our importance sampling scheme (Our IS) and the adaptive sampling variant (Our AS) can improve over classical SGD, LOW~\cite{santiago2021low} and DLIS~\cite{katharopoulos2018dlis} on modern transformer architecture.
    }
    \label{fig:convergence_plots_CIFAR10_tranformer_appendix}
    
\end{figure}

\begin{figure}[h]
    \centering
    \resizebox{\columnwidth}{!}{
    \begin{tikzpicture}
        \node[inner sep=0pt] (russell) at (6.88,2.9)
        {\includegraphics[scale=0.31]{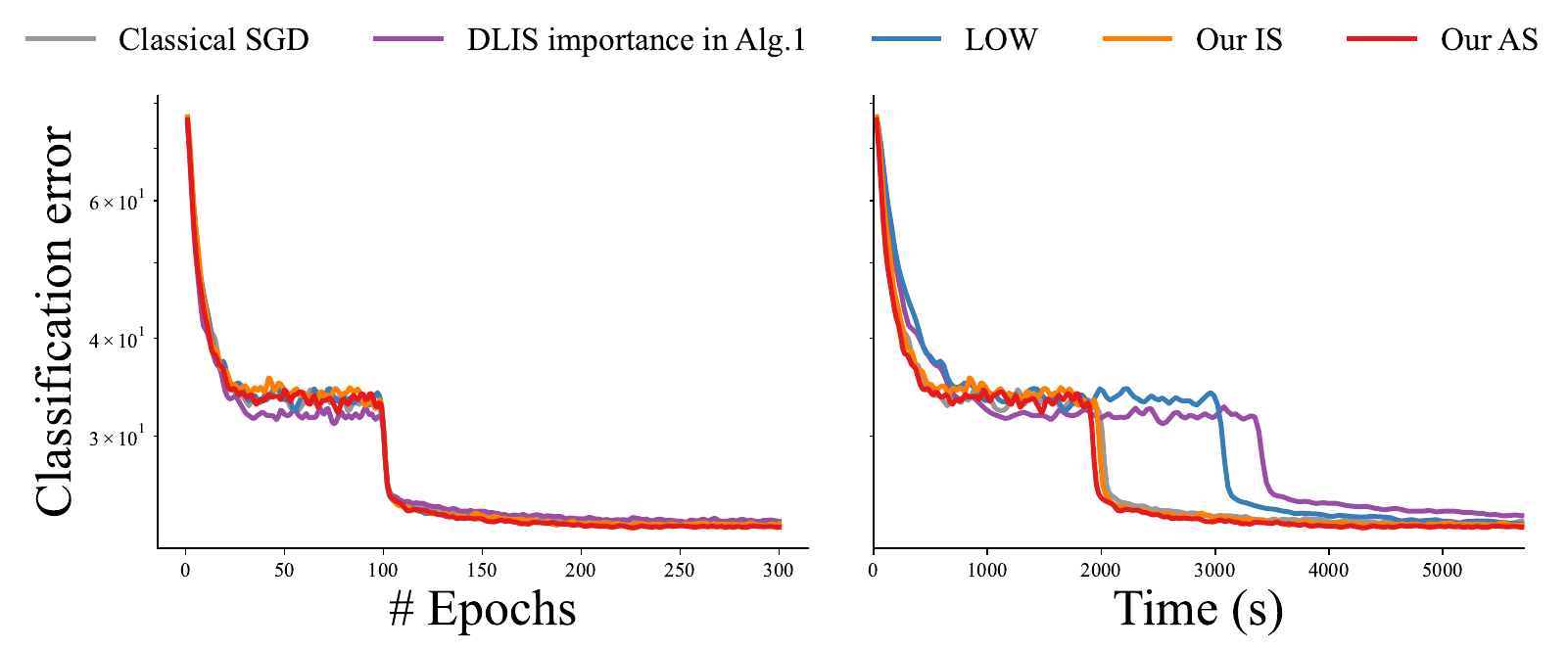}};
        \node[inner sep=0pt] (russell) at (10,3.4)
        {\includegraphics[height=0.9cm]{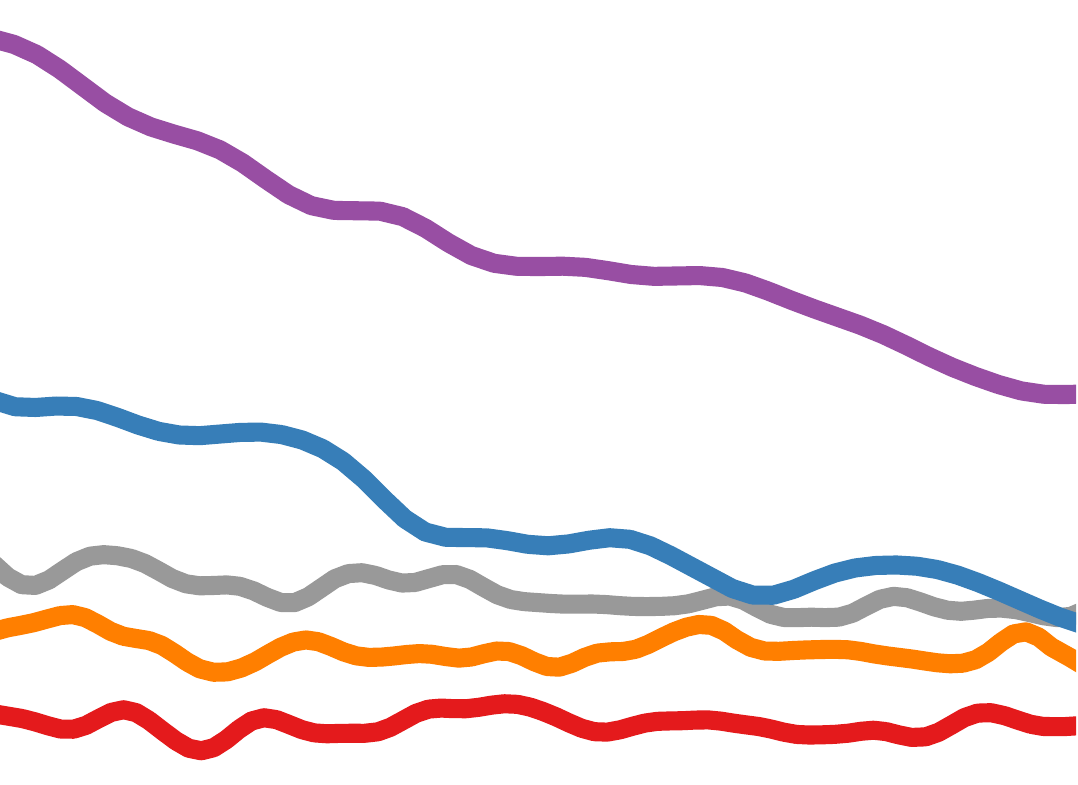}};
        \begin{scope}
            \draw[black,thick] (9.4,2.82) -- (9.4,3.8) -- (10.6,3.8) -- (10.6,2.82) -- cycle;
        \end{scope}
        \draw[black,thick] (9.5,1.75) rectangle (10.8,2.0);
        \draw[black,thick] (10.8,2.0) -- (10.6,2.82);
        \draw[black,thick] (9.5,2.0) -- (9.4,2.82);
        \node[inner sep=0pt] (russell) at (6.,3.3)
        {\includegraphics[height=0.9cm]{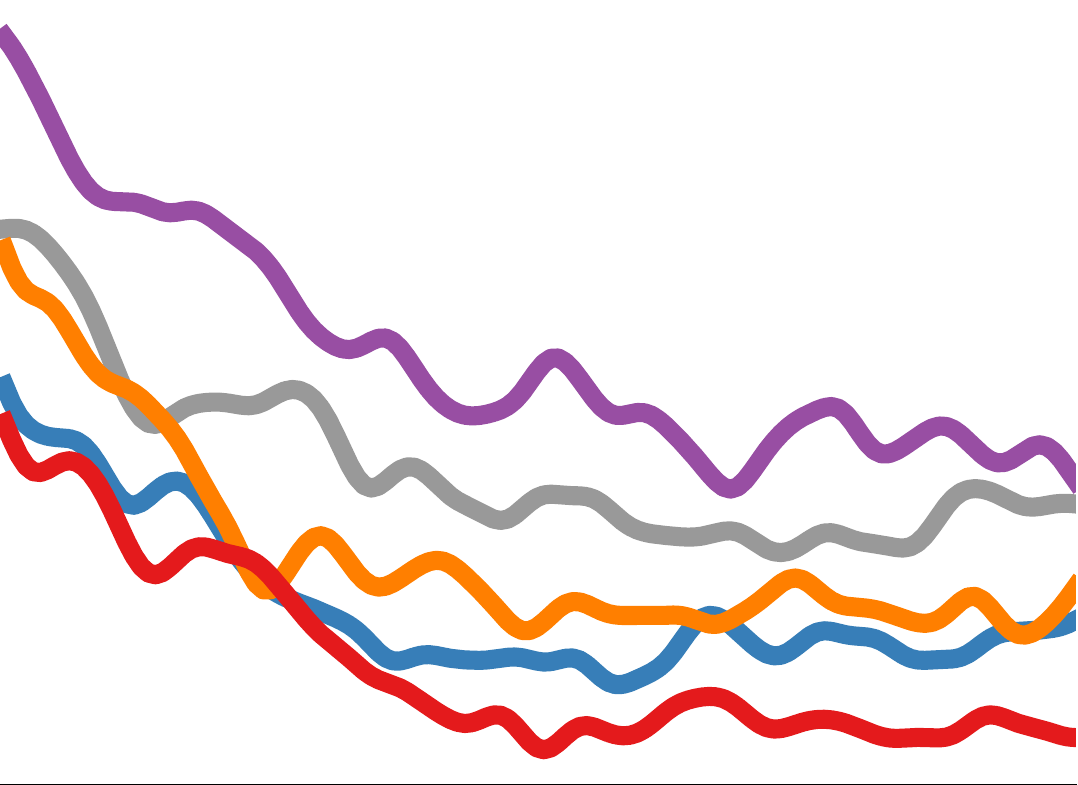}};
        \begin{scope}
            \draw[black,thick] (5.4,2.82) -- (5.4,3.8) -- (6.6,3.8) -- (6.6,2.82) -- cycle;
        \end{scope}
        \draw[black,thick] (5.5,1.75) rectangle (6.8,2.0);
        \draw[black,thick] (6.8,2.0) -- (6.6,2.82);
        \draw[black,thick] (5.5,2.0) -- (5.4,2.82);
    \end{tikzpicture}
    }
    \caption{
        On CIFAR-100 classification dataset, instead of comparing the DLIS resampling algorithm, we use DLIS importance in our \cref{alg:IS}. We display zoom-in of the end of the curves to highlight the differences.
        At equal epochs (left), our methods (Our IS \& Our AS) show improvements compared to LOW~\cite{santiago2021low} and DLIS weights. 
        At equal time (right), LOW and the DLIS weights takes longer to converge. Overall our approach shows faster convergence with lower importance computation.
    }
    \label{fig:convergence_plots_CIFAR100_DLIS_LOW_equal_time_appendix}
\end{figure}

\Cref{fig:convergence_plots_algo_comp_DLIS_MINIS_loss,fig:convergence_plots_CIFAR10_tranformer_appendix,fig:convergence_plots_CIFAR100_DLIS_LOW_equal_time_appendix} show that our adaptive sampling variant (our AS) can achieve better results than our IS or our OMIS in practice.
Our AS is a heuristic and we leave its theoretical formulation as future work.

\end{document}